\documentclass[preprint,12pt]{elsarticle}
\usepackage{times}
\usepackage{caption}
\usepackage{graphicx}
\usepackage{amsmath}
\usepackage{amsthm}
\usepackage{booktabs}
\usepackage{algorithm}
\usepackage{algorithmic}
\usepackage{graphicx}
\usepackage{amsmath}
\usepackage{hyperref}
\usepackage{amssymb}
\usepackage{booktabs}
\usepackage{multirow}
\usepackage[capitalize]{cleveref}
\crefname{section}{Sec.}{Secs.}
\Crefname{section}{Section}{Sections}
\Crefname{table}{Table}{Tables}
\crefname{table}{Tab.}{Tabs.}
\usepackage{bbding}
\usepackage{overpic}
\usepackage{subfigure}
\usepackage{xcolor}

\DeclareRobustCommand\onedot{\futurelet\@let@token\@onedot}
\def\eg{\emph{e.g., }}  

\def\ie{\emph{i.e., }}

\def\etal{\emph{et al.} }
\makeatother

\graphicspath{{Figures/}}

\journal{Pattern Recognition}

\begin{document}
\begin{frontmatter}
\title{Restoring Vision in Hazy Weather with Hierarchical Contrastive Learning}

\author[1]{Tao Wang}
\ead{taowangzj@gmail.com}
\author[1]{Guangpin Tao}
\ead{tgpin@smail.nju.edu.cn}
\author[2]{Wanglong Lu}
\ead{wanglongl@mun.ca}
\author[3]{Kaihao Zhang}
\ead{super.khzhang@gmail.com}
\author[4]{Wenhan Luo}
\ead{whluo.china@gmail.com}
\author[5]{Xiaoqin Zhang}
\ead{zhangxiaoqinnan@gmail.com}
\author[1]{Tong Lu\corref{cor1}}
\ead{lutong@nju.edu.cn}
\cortext[cor1]{Corresponding author}
\affiliation[1]{organization={National Key Lab for Novel Software Technology, Nanjing University},
            city={Nanjing},
            country={China}}
\affiliation[2]{organization={Memorial University of Newfoundland},
            city={St.John’s},
            country={Canada}}
\affiliation[3]{organization={Australian National University},
            city = {Canberra},
            country={Australia}}
\affiliation[4]{organization={Sun Yat-sen University},
            city = {Shenzhen},
            country={China}}
\affiliation[5]{organization={Wenzhou University},
            city = {Wenzhou},
            country={China}}

\begin{abstract}
Image restoration under hazy weather condition, which is called single image dehazing, has been of significant interest for various computer vision applications. 
In recent years, deep learning-based methods have achieved success. However, existing image dehazing methods typically neglect the hierarchy of features in the neural network and fail to exploit their relationships fully. To this end, we propose an effective image dehazing method named Hierarchical Contrastive Dehazing (HCD), which is based on feature fusion and contrastive learning strategies. HCD consists of a hierarchical dehazing network (HDN) and a novel hierarchical contrastive loss (HCL). Specifically, the core design in the HDN is a hierarchical interaction module, which utilizes multi-scale activation to revise the feature responses hierarchically. To cooperate with the training of HDN, we propose HCL which performs contrastive learning on hierarchically paired exemplars, facilitating haze removal. Extensive experiments on public datasets, RESIDE, HazeRD, and DENSE-HAZE, demonstrate that HCD quantitatively outperforms the state-of-the-art methods in terms of PSNR, SSIM and achieves better visual quality.
\end{abstract}


\end{frontmatter}

\section{Introduction}\label{section:Introduction}
Single image dehazing aims to recover the latent haze-free image from a given hazy image. Due to its wide range of applications (\eg autonomous driving and video surveillance), single image dehazing has become a hot topic in the fields of computer vision and image processing.

Traditional image dehazing methods~\cite{ali2023boundary} are mostly based on the image prior and the atmosphere scattering model (ASM) \cite{tan2008visibility}. Specifically, the pipeline of these methods is to find some prior information from images to estimate the transmission map $t$ and global atmosphere light $A$ from the haze image $I$, and then to use the predicted $t$ and $A$ to recover the clear image $J$ according to ASM as $J(x) = (I(x)-A)/t(x)+A$, where $x$ is the pixel position. Unfortunately, traditional methods usually require time-consuming iteration optimization and handcrafted priors. Thus they may not work well in complex haze scenarios.

In recent years, with the rapid development of deep learning techniques and the collection of large-scale synthetic datasets, many data-driven image dehazing approaches have been proposed to achieve haze removal. In the beginning, many works like \cite{li2017aod,cai2016dehazenet} attempt to estimate the transmission map and the atmospheric light through Convolution Neural Networks (CNNs) and then restore the clear image via ASM. However, the inaccurate estimation of the transmission map or atmospheric light may easily lead to their poor dehazing performance. More recently, another class of data-driven approaches \cite{li2022dual,lin2022msaff} directly ignores ASM and uses an end-to-end CNN to learn a mapping between the hazy image and the clear image. For example, Jiang \etal \cite{jiang2023deep} design an end-to-end network containing a haze residual attention sub-network and a detail refinement sub-network to directly recover the clear image from hazy input. The Attention mechanism~\cite{liu2023local,sun2023multi} is embedded into the end-to-end network for effective image dehazing.
Even though the above data-driven methods greatly improve the visual quality of dehazed results, they share the following drawbacks: 
1) \emph{They do not fully exploit hierarchical features in CNNs.} As we know, shallow features of CNNs contain more details and spatial information, while deep features focus on higher-level context and semantic information \cite{hariharan2015hypercolumns}. Both shallow and deep features of CNNs are beneficial for the process of image dehazing. However, existing methods \cite{qin2020ffa,qu2019enhanced} do not fully exploit complementary information from these hierarchical features in CNNs, and it is easy to cause color distortion in the recovered images~\cite{zhang2020multi}.
2) \emph{They only consider positive-oriented supervision information in the training stage.} Most data-driven image dehazing methods typically regard haze-free images as positive samples to guide the optimization of the model and do not fully mine the hazy input images (negative samples) in the training stage. Ignoring negative-oriented learning reduces the representation ability of the model to some extent, which results in lower image restoration performance of the model~\cite{wu2021contrastive}.

\begin{figure*}[t]
	\centering
	 \begin{overpic}[width=0.9\textwidth]{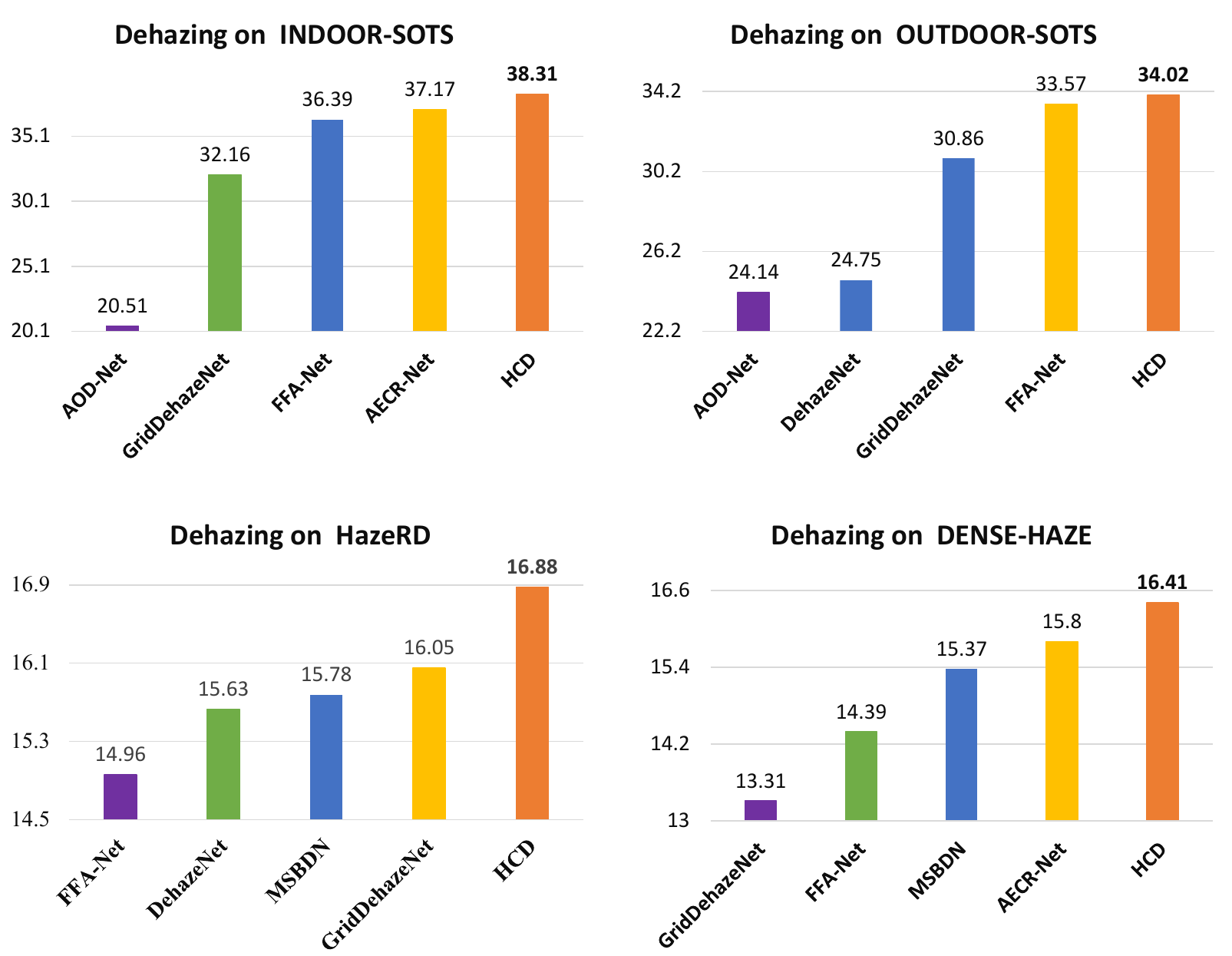}
 \end{overpic} 
 
 \caption{Performance comparison of the proposed HCD with the state-of-the-art methods on popular image dehazing datasets. Our HCD significantly advances the state-of-the-art image dehazing performance in terms of PSNR. 1) +~\textbf{1.14} dB on the indoor subset of Synthetic Objective Testing Set (SOTS)~\cite{li2018benchmarking}, 2) +~\textbf{0.45} dB on the outdoor subset of SOTS~\cite{li2018benchmarking}, 3) +~\textbf{0.83} dB on the HazeRD dataset~\cite{zhang2017hazerd}, and 4) +~\textbf{0.61} dB on the DENSE-HAZE dataset~\cite{zhang2017hazerd}.} 
	\label{fig:introduction}
\end{figure*}
To address the problems mentioned above, we aim to design an effective feature fusion scheme in the network and a positive and negative-oriented supervision strategy to further supervise the network training. To this end, we propose a novel dehazing method called Hierarchical Contrastive Dehazing (HCD). The proposed HCD consists of a hierarchical dehazing network (HDN) and a hierarchical contrastive loss (HCL). Specifically, HDN includes a hierarchical feature extractor, a hierarchical interaction module (HIM), and a multi-output image reconstruction module. The hierarchical feature extractor extracts hierarchical features (\ie multi-resolution features) from a hazy input. The features are then fed into the HIM, which allows information
flow to exchange efficiently across different branches and
improves the dehazing performance. This module first progressively propagates higher-level features into the shallow layers to suppress the noise in lower-level features and then incorporates the detailed information from lower layers into the deep layers. After that, the multi-output image reconstruction module assembles features with three resolutions and reconstructs clean images. Finally, HCL is embedded into our special hierarchical structure dehazing network. It guides the network to exploit a representation by maximizing similarity and dissimilarity over samples that are organized into similar and dissimilar pairs in a hierarchical manner. 

The closest image dehazing methods with our proposed HCD are GridDehazeNet~\cite{liu2019griddehazenet} and AECR-Net~\cite{wu2021contrastive}. However, our HCD differs from GridDehazeNet and AECR-Net in several ways. First, our proposed HFB refines features by using differences between features in different branches and hierarchically propagating information from the bottom to the top branch, allowing it to effectively utilize the hierarchical features in the network. In contrast, GridDehazeNet and AECR-Net use channel-wise attention and skip connections, which cannot fully utilize hierarchical features in the network. Second, our proposed hierarchical contrastive loss is applied at multiple scales using negative and positive samples to enhance the feature representation ability of the network. However, GridDehazeNet only considers the information of positive images as an upper bound, and AECR-Net only performs contrastive learning on a single image scale. By employing a hierarchical structure for contrastive learning, we further enhance the network's feature representation ability. Finally, our extensive experiments demonstrate that the proposed HCD significantly outperforms state-of-the-art image dehazing methods, including GridDehazeNet and AECR-Net, as illustrated in Fig. \ref{fig:introduction}.

To summarize, the contributions of our work are as follows:
 \begin{itemize}
 \item We propose a novel Hierarchical Contrastive Dehazing (HCD) method, which employs a hierarchical feature fusion technique and a contrastive learning strategy to effectively enhance the feature representation ability of the model. 
 \item The implementation of a hierarchical interaction module in a hierarchical structure network allows information flow to exchange efficiently across different branches and improves the dehazing performance.
 
 \item By considering both positive and negative-oriented supervision, the proposed hierarchical contrastive loss effectively guides the model to learn the valid features for the image dehazing. 
 
 \item Extensive experimental results on benchmarks demonstrate that HCD performs favorably against state-of-the-art approaches.
 \end{itemize}

The remainder of this paper is organized as follows. Sec. \ref{sec:related_work} presents the related work. Sec. \ref{sec:method} introduces our proposed method. Sec. \ref{sec:experiment} reports experimental results. Sec. \ref{sec:conclusion} provides a conclusion of this paper.

\section{Related Work}\label{sec:related_work}
The proposed method is related to image dehazing and contrastive learning, which are reviewed in the following.

\subsection{Single Image Dehazing} 
Image dehazing aims to recover a clear image from the hazy image, which is a popular research topic in the computer vision community. A wide range of methods has been proposed in the literature to address this problem. They are approximately categorized into traditional prior-based methods and deep learning-based methods.

Traditional prior-based methods mainly focus on exploring the statistical properties of images (\ie statistical prior) to estimate the atmospheric light and transmission map and then recover the clear image by ASM~\cite{mutimbu2018factor}. Tan \cite{tan2008visibility} proposes an image dehazing method by maximizing the local contrast of hazy images, which is based on the statistical observation that clear images have more contrast than hazy images. He \etal \cite{he2010single} propose a haze removal approach utilizing the dark channel prior. This prior is motivated by the assumption that the dark channels of clear images are close to zero. In \cite{fattal2014dehazing}, the color-line prior is employed to achieve image dehazing. The color-line prior hypothesises that pixels in small image patches have the characteristic of a one-dimensional distribution in RGB space. Zhu \etal \cite{zhu2015fast} employ a linear model to estimate the depth information of images based on the color attenuation prior for image haze removal. Yuan \etal \cite{yuan2021confidence} propose a new unified framework for image dehazing, which aims to effectively use  several existing priors to obtain clear images from the hazy image. Although the prior-based methods have achieved impressive results, the representation ability of these hand-crafted priors is limited, especially for highly complex hazy scenes.

In recent years, with the rapid development of deep learning, deep learning-based fog removal methods have been extensively studied. Deep learning-based dehazing methods can be approximately divided into supervised methods, semi-supervised methods, and unsupervised methods~\cite{gui2022comprehensive}. For supervised dehazing methods, some methods are designed based on ASM model. For instance,  Cai \etal \cite{cai2016dehazenet} first employ a convolutional neural network to estimate the transmission map from hazy images and then restore dehazed images based on the atmospheric scattering model. 
Li \etal \cite{li2017aod} first reformulate the scattering model, which translates the problem of estimating the transmission and atmospheric light into estimating  an intermediate parameter. Then, based on the re-formulated scattering model, they propose an AOD-Net model to achieve image dehazing. On the other hand, some approaches directly learn the mapping from hazy and clear images through convolutional neural networks to achieve image dehazing.
For example, Qin \etal \cite{qin2020ffa} design a deep FFA-Net for the dehazing task. The core component in the FFA-Net is the feature attention module, which includes a pixel attention block, a channel attention block, and a residual operation. In \cite{yin2020novel}, Yin \etal propose a parallel attention network for image dehazing, which mainly designs a parallel spatial/channel-wise attention block to capture more informative spatial and channel-wise features. As for semi-supervised dehazing methods, representative approaches such as PSD~\cite{chen2021psd} and SSDT~\cite{zhang2021single} first utilize a backbone network for pre-training purposes to acquire a basic network that fits synthetic data. Then, the unsupervised fine-tuning process on real-world domains is applied to the network to improve the capability of the network to deal with real word hazy images. The supervised and semi-supervised dehazing methods rely on paired data in the training process, which limits their application. Therefore, some unsupervised dehazing methods are proposed. For example, inspired by CycleGAN~\cite{zhu2017unpaired}, Cycle-Dehaze~\cite{engin2018cycle} and CDNet~\cite{dudhane2019cdnet} achieve image dehazing by unsupervised domain translation.

\subsection{Contrastive Learning}
Recently, contrastive learning has been widely used in self-supervised representation learning. The goal of contrastive learning is to learn an invariant representation from the data in the training dataset. The key step of the contrastive learning technique is to design an effective strategy to maximize the complementary information over data samples. Some contrastive learning methods improve representation ability by designing a contrastive loss, such as triplet loss and InfoNCE loss. The contrastive loss is used to push an exemplar close to similar samples while pushing it far away from dissimilar samples. For example, Park \etal \cite{park2020contrastive} use the InfoNCE loss to train the network for unpaired image-to-image translation and demonstrate that contrastive learning techniques can significantly improve the performance of models in conditional image synthesis tasks. With the rapid development of contrastive learning techniques, several low-level vision tasks have employed contrastive loss and achieved promising performance. Zhang \etal \cite{zhang2021blind} employ the contrastive learning technique to solve blind super-resolution in real-world scenery. They design contrastive decoupling encoding for learning resolution-invariant features and use these learned features to obtain high-resolution images. Recently, Wu \etal \cite{wu2021contrastive} propose a novel pixel-wise contrastive loss and regard it as a regularization term to train a network for image dehazing. Unlike Wu \etal \cite{wu2021contrastive} only perform contrastive learning on a single image scale, we employ a hierarchical structure for contrastive learning to further enhance the network's feature representation ability, resulting in a significant improvement in haze removal performance.

\begin{figure*}[t]
\begin{center}
	\includegraphics[width=\textwidth]{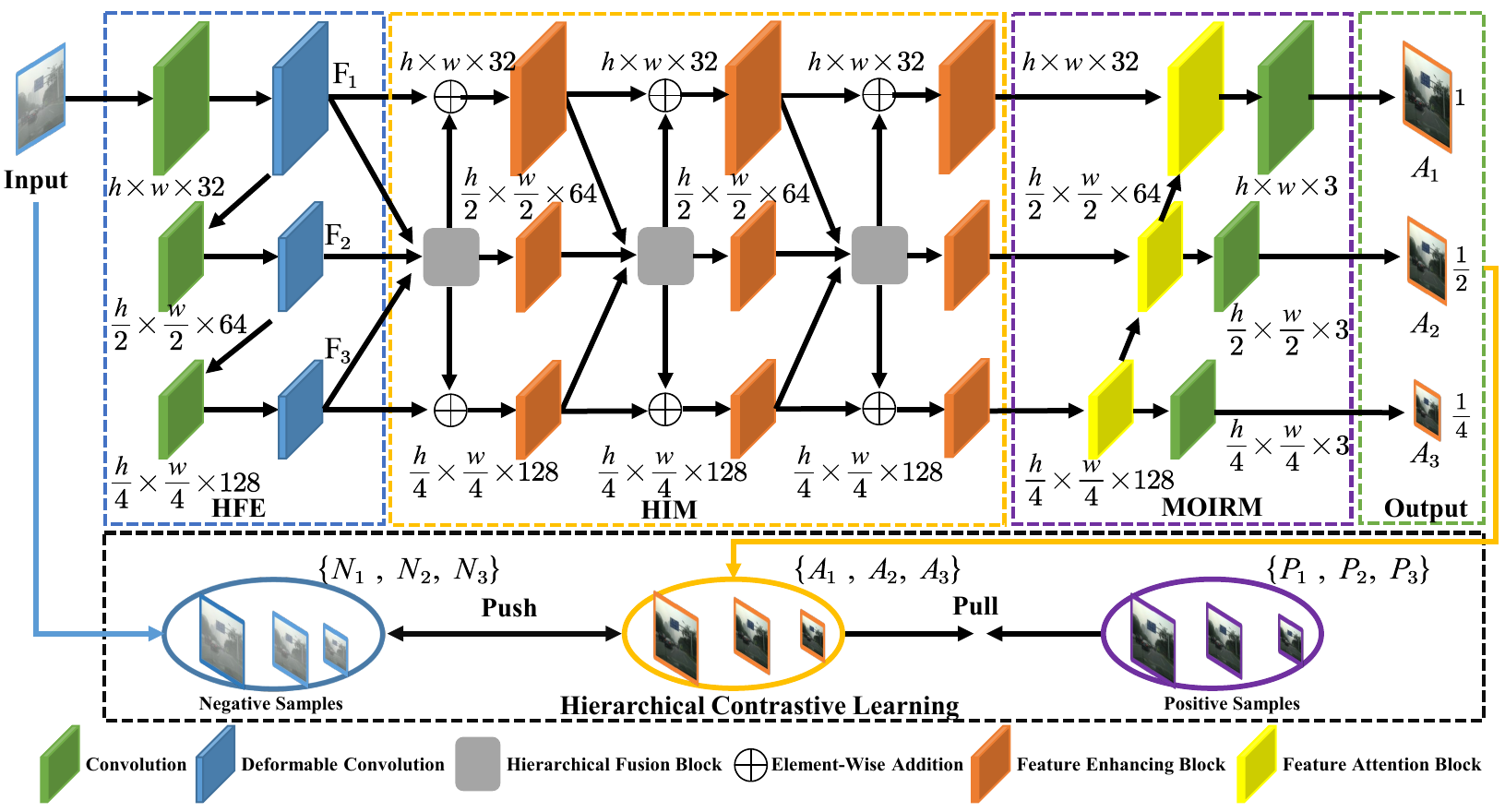}
	\caption{The architecture of our HCD. It includes a hierarchical dehazing network (HDN) shown at the top and a hierarchical contrastive loss (HCL) shown at the bottom. HDN consists of a hierarchical feature extractor (HFE), a hierarchical interaction module (HIM), and a multi-output image reconstruction module (MOIRM). In HIM, the hierarchical fusion block(HFB) is the core component, which is shown in Fig.~\ref{fig:module}. The proposed HCD employs contrastive constraints in a hierarchical structure manner to perform feature representation learning, which can better help haze removal. 
	}
	\label{fig:overall}
 \end{center}
\end{figure*}
\section{Proposed Method}\label{sec:method}
In this section, we first introduce an overview of the proposed HCD and then detail each component within it respectively. The loss function to optimize the network is introduced in the end. 

\subsection{Method Overview}\label{sec:overview}
We propose the HCD method for the image dehazing task, which can fully exploit hierarchical representation from the hazy image. The overall architecture of HCD is illustrated in Fig. \ref{fig:overall}. Our HCD includes a hierarchical dehazing network (HDN) and a hierarchical contrastive loss (HCL). HDN is used to dehaze the input image, and HCL utilizes the information of positive and negative samples to guide the HDN training. These two components are working together to produce a good performance in image dehazing. Specifically, as shown in Fig. \ref{fig:overall}, given a hazy input image, a hierarchical feature extractor (HFE) first extracts hierarchical visual features, then a hierarchical interaction module (HIM) fuses these hierarchical features alternately and hierarchically. After that, a multi-output image reconstruction module (MOIRM) reconstructs the output features of HIM and generates different multi-scale dehazed images. Finally, HDN performs contrastive learning in a hierarchical manner via HCL to further improve the feature learning ability. In the following subsections, we detail each component of our HCD, \ie HDN and HCL.

\subsection{Hierarchical Dehazing Network}\label{sec:structure1}
\textbf{Hierarchical Feature Extractor.} Feature representation plays an essential role in the computer vision community, and the representation directly affects the performance of a deep learning method~\cite{bengio2013representation}. As discussed in \cite{bengio2013representation}, a good feature representation should have the following characteristics. One is that it can capture multiple configurations from the input. Another is that it should organize the explanatory factors of the input data as a hierarchy, where more abstract concepts are at a higher level. To this end, we propose a hierarchical feature extractor (HFE) in the network to effectively extract features from a hazy input image.

HFE module is designed to extract multi-scale and hierarchical features from input images, which are then used for subsequent feature fusion and dehazing. In HFE, we improve the feature extraction capability of the network via deformable convolution to expand the receptive field with an adaptive shape. Specifically, as shown in Fig. \ref{fig:overall}, HFE is designed under three parallel branches to produce hierarchical features with different resolutions and depths. Each branch in HFE is composed of a $3\times 3$ convolution $\operatorname{Conv}$ and a deformable convolution $\operatorname{DCN}$ \cite{dai2017deformable}. The convolution is used to transform the resolution and depth of the input feature, and the deformable convolution is employed to extract abundant features. We experimentally demonstrate that $\operatorname{DCN}$ increases the dehazing performance of the model. In particular, for a hazy input image $N$, the output feature resolution of the upper branch is the same as $N$, and for the other two branches, the resolutions of features are decreased by  factors of $\frac{1}{2}$ and $\frac{1}{4}$ respectively. In addition, the depths of hierarchical features $\mathbf{F_{1}},\mathbf{F_{2}},\mathbf{F_{3}}$ outputted from the three branches are $32$, $64$, and $128$, respectively. 
 
\textbf{Hierarchical  Interaction Module.} As discussed in previous work, conventional networks are susceptible to bottleneck effects \cite{liu2019griddehazenet} because the progressive down-sampling operations in the feature extractor stage cause the feature information loss problem. Therefore, we propose a plug-and-play hierarchical interaction module (HIM), as shown in Fig. \ref{fig:overall}, to let information flow exchange effectively across different branches in the network. HIM is located between the hierarchical feature extractor and the multi-output image reconstruction module, which contains three identical sub-modules. More specifically, each sub-module consists of a hierarchical fusion block (HFB) and three feature enhancement blocks (FEB). The features $\{\mathbf{F_{1}},\mathbf{F_{2}},\mathbf{F_{3}}\}$ from HFE are firstly enhanced by HFB and are then refined by FEB. For simplicity,  we introduce one sub-module in the following.

The input hierarchical features $\{\mathbf{F_{1}},\mathbf{F_{2}},\mathbf{F_{3}}\}$ are first processed by HFB. As shown in Fig. \ref{fig:module}, to fully exploit hierarchical features from non-adjacent stages, HFB is divided into two core steps: 
 \begin{figure*}[t]
	\centering
	\includegraphics[width=\textwidth]{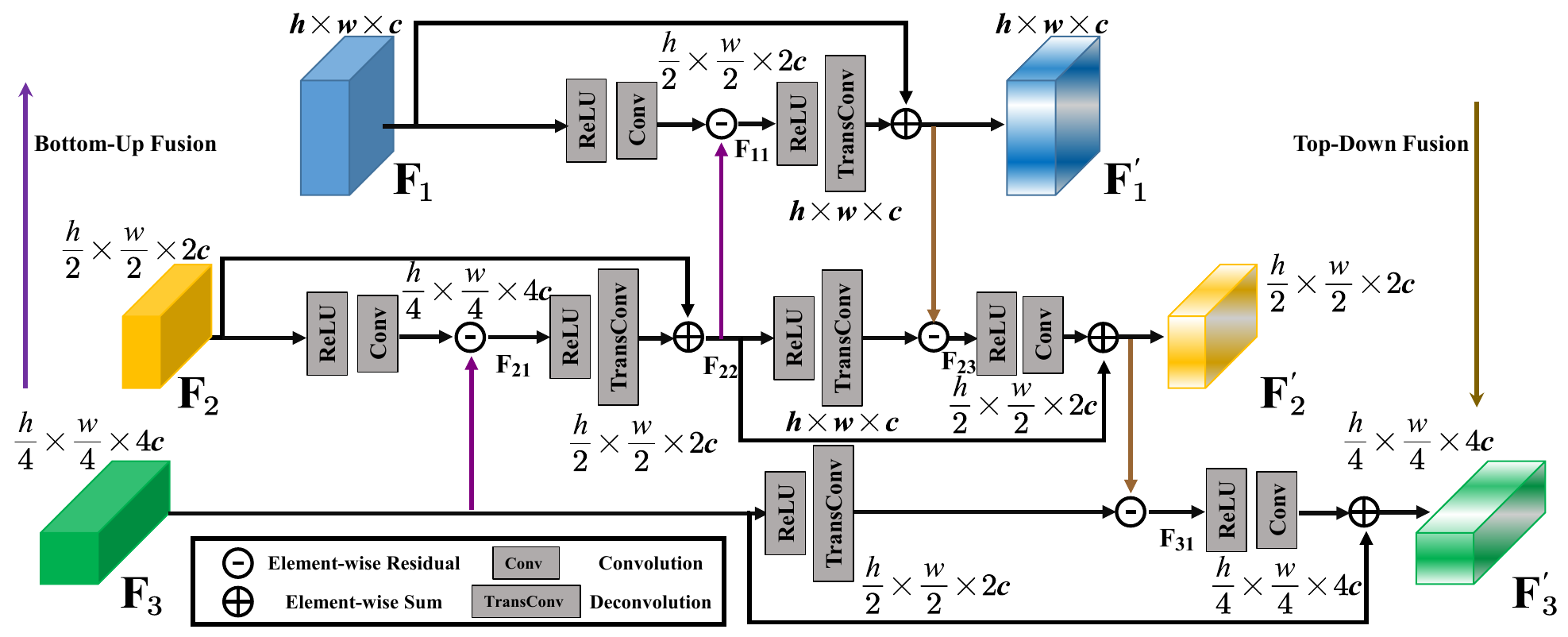}
	\caption{Illustration of the hierarchical fusion block (HFB). It contains two steps: Bottom-Up Fusion and Top-Down Fusion. $\mathbf{F_{1}},\mathbf{F_{2}},\mathbf{F_{3}}$ are input hierarchical features, and $\mathbf{F_{1}^{'}},\mathbf{F_{2}^{'}},\mathbf{F_{3}^{'}}$ are corresponding updated versions by HFB. Element-wise Residual operation refers to obtaining the residual feature by subtracting the corresponding elements of two features. The Bottom-Up Fusion and Top-Down Fusion stages can be formulated as in Eq.~\ref{top_fusion} and Eq.~\ref{down_fusion} respectively. }
	\label{fig:module}
\end{figure*}
 \begin{figure*}[t]
	\centering
	\includegraphics[width=\textwidth]{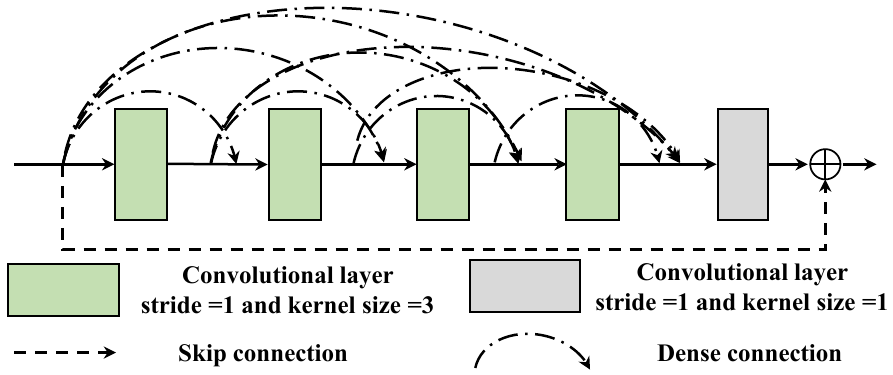}
	\caption{Illustration of the feature enhancement block (FEB). FEB block consists of five densely connected convolutional layers and a residual skip connection.}
	\label{fig:FEB}
\end{figure*}
\textit{Bottom-Up Fusion:} HFB firstly hierarchically propagates information from the bottom branch to the top branch. This process contains two fusion stages. In the first stage, feature $\mathbf{F_{2}}$ from the second branch is progressively forwarded by a $\operatorname{ReLU}$ non-linear activation function and a $3 \times 3$ convolution to produce the enhanced feature $F_{21}$. Inspired by \cite{dong2020multi}, we then compute the difference between $F_{21}$ and $\mathbf{F_{3}}$ and update the feature $\mathbf{F_{2}}$ with the computed difference. Again, in the second stage, feature $\mathbf{F_{1}}$ from the top branch is processed by a $\operatorname{ReLU}$ and a convolution. We then compute the difference between the output of the first stage and $\mathbf{F_{1}}$, and obtain the updated feature $\mathbf{F_{1}^{'}}$. In this way, the high-level context information from the bottom branch is propagated to the top branch. This process is formulated as:
\begin{equation}
\begin{aligned}
F_{21}&=\mathbf{F_{3}} - \operatorname{Conv}(\operatorname{ReLU}(\mathbf{F_{2}})),\\
F_{22} &= \operatorname{TransConv}(\operatorname{ReLU}(F_{21})) + \mathbf{F_{2}},\\
F_{11}&=F_{22} - \operatorname{Conv}(\operatorname{ReLU}(\mathbf{F_{1}})),\\
\mathbf{F_{1}^{'}} &= \operatorname{TransConv}(\operatorname{ReLU}(F_{11})) + \mathbf{F_{1}},\\
\end{aligned}
\label{top_fusion}
\end{equation}
where $\mathbf{F_{1}^{'}}$ is the updated version of $\mathbf{F_{1}}$, $\operatorname{Conv}$ refers to a $ 3 \times 3$ convolution with a stride of $2$, and $\operatorname{TransConv}$ denotes deconvolution that is applied to transform the shapes of features so that the features from different scales can be used.

\textit{Top-Down Fusion:} To further fuse the hierarchical features, we design a symmetric hierarchical top-down fusion structure in HFB. As illustrated in Fig. \ref{fig:module}, similar to hierarchical bottom-up fusion, hierarchical top-down fusion has two fusion stages. In the first fusion stage, a ReLU activation function and a deconvolution are employed to transform shapes of feature $F_{22}$ to the same as $\mathbf{F_{1}^{'}}$. Then, the differences between $F_{22}$ and $\mathbf{F_{1}^{'}}$ are used to refine the feature $F_{22}$. After that, in the second fusion stage, the refined feature $\mathbf{F_{2}^{'}}$ is fed into the bottom branch to further refine $\mathbf{F_{3}}$. The process of the top-down fusion can be presented as:
\begin{equation}
\begin{aligned}
F_{23}&= \mathbf{F_{1}^{'}} - \operatorname{TransConv}(\operatorname{ReLU}(F_{22})),\\
\mathbf{F_{2}^{'}} &= \operatorname{Conv}(\operatorname{ReLU}(F_{23})) + F_{22},\\
F_{31}&=\mathbf{F_{2}^{'}} - \operatorname{TransConv}(\operatorname{ReLU}(\mathbf{F_{3}})),\\
\mathbf{F_{3}^{'}} &= \operatorname{Conv}(\operatorname{ReLU}(F_{31})) + \mathbf{F_{3}},\\
\end{aligned}
\label{down_fusion}
\end{equation}
where $\mathbf{F_{1}^{'}},\mathbf{F_{2}^{'}},\mathbf{F_{3}^{'}}$ are the outputs of HFB. In the end, as shown in Fig. \ref{fig:overall},  the outputs of HFB are further strengthened via parallel residual connection and FEB, where FEB is the Residual Dense Block in \cite{liu2019griddehazenet}. As shown in Fig.~\ref{fig:FEB}, FEB block consists of five densely connected convolutional layers and a residual skip connection. The first four layers use a kernel size of 3 to increase the number of feature maps. Subsequently, the last layer utilizes a kernel size of 1 to fuse these feature maps. Finally, a skip connection is employed to combine the output of this block with its input. Our HIM module can thus exploit multi-scale hierarchical features to improve the dehazing performance.

\textbf{Multi-output Image Reconstruction Module.} In HIE, different output branches produce feature maps with different resolutions. We consider that these hierarchical feature maps with different characteristics can be used to produce different samples for subsequent contrastive learning. Thus, we design a multi-output image reconstruction module in the tail of the network. As illustrated in Fig. \ref{fig:overall}, in each branch, we first apply a feature attention block (FAB) to refine the feature, and then use a single convolution layer to reconstruct the image. The image reconstruction in each branch can be formulated as follows: 
\begin{equation}
A_{n}= \begin{cases}\operatorname{Conv}\left(\mathrm{FAB}_{n}\left(\left(\mathrm{FAB}_{n+1}^{\text{out}}\right)^{\uparrow}; \mathrm{HIM}_{n}^{\text {out}}\right)\right), & n=1,2 \\ \operatorname{Conv}\left(\mathrm{FAB}_{n}\left(\mathrm{HIM}_{n+1}^{\text {out}} \right)\right), & n=3\end{cases}
\end{equation}
where $\mathrm{HIM}_{n}^{\text {out}}$, $\mathrm{FAB}_{n}^{\text{out}}$ are the outputs of the $n^{th}$ branch HIM and FAB. $\operatorname{Conv}$ is a $3\times3$ convolution. Up-sampling $\uparrow$ is used such that the features from different scales can be fused.

 \begin{figure*}[t]
	\centering
	\includegraphics[width=0.85\textwidth]{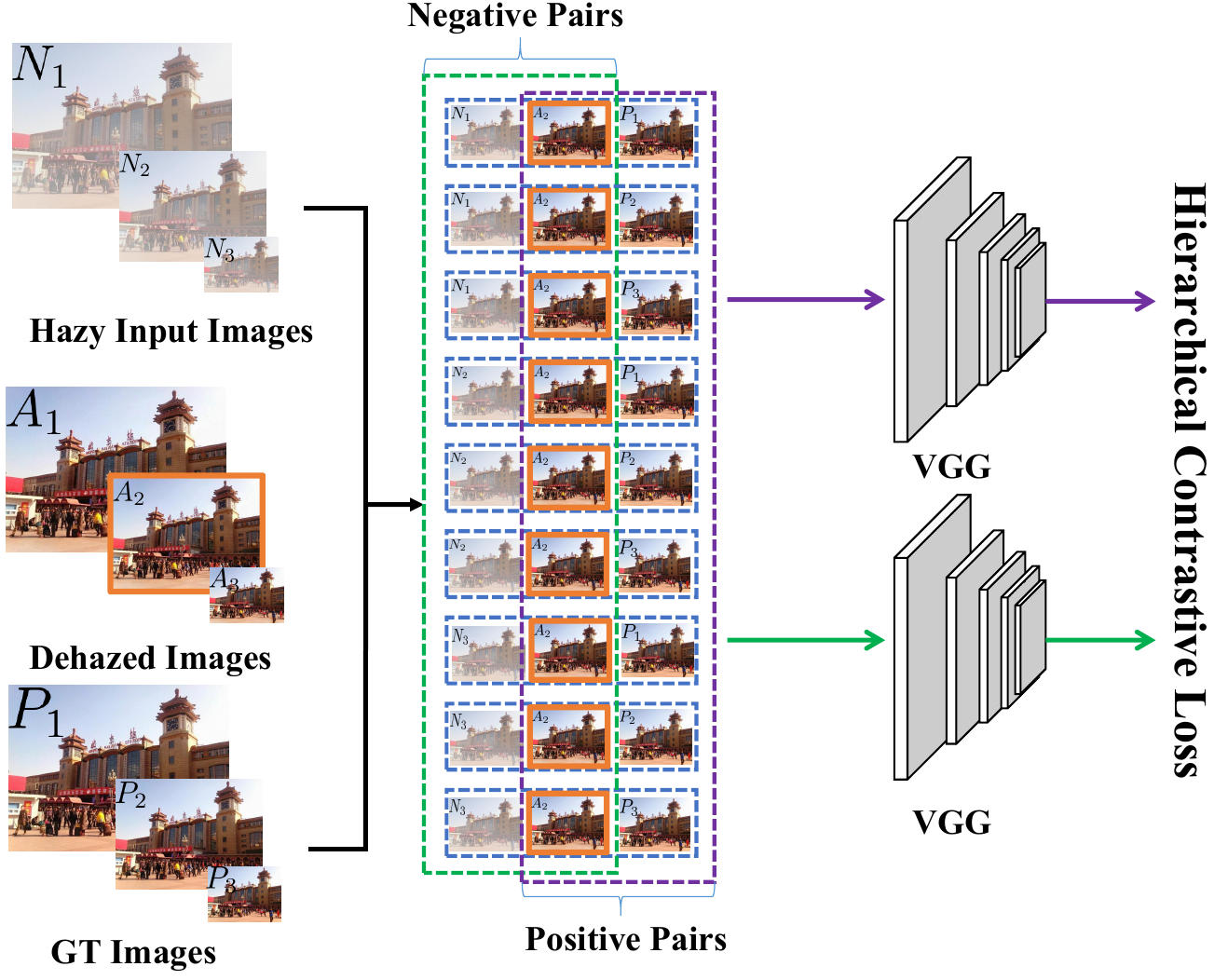}
	\caption{Illustration of the proposed hierarchical contrastive learning. Taking the recovered image $A_{2}$ as an example, we construct different types of positive and negative pairs. The image $A_{2}$, positive samples and negative samples are encoded into features by a pre-trained VGG-19 network. The hierarchical contrastive loss compares these features and guides the network to learn more useful information for image dehazing.}
   \label{fig:hcl}
\end{figure*}

\subsection{Hierarchical Contrastive Learning}\label{sec:structure4}
The proposed HDN recovers the images in a hierarchical structure. Therefore, it is natural that we consider joint optimization of all the dehazed images output by HDN. To this end, we resort to contrastive learning, which is a discriminant-based technique that pulls similar samples closer while pushing dissimilar samples away \cite{park2020contrastive,wu2021contrastive}. Though contrastive learning has shown effectiveness in many high-level vision tasks, its potential for single image dehazing has not been fully explored. We thus propose a novel hierarchical contrastive loss to guide our HDN to remove complex haze components. Two aspects are required to consider when constructing our loss: one is to build the appropriate positive and negative pairs of samples, and the other is to find a suitable latent space to differentiate sample distribution. For the first aspect, benefiting from the hierarchical structure of HDN, we can collect abundant contrastive samples in different resolutions. We label hazy input images in three resolutions as negative samples, and regard their corresponding ground truth images as positive samples. For the second aspect, inspired by~\cite{wu2021contrastive}, we employ a pre-trained VGG-19 network~\cite{simonyan2014very} to obtain feature embedding for measuring feature similarity. The VGG-19 network~\cite{simonyan2014very} is trained on ImageNet dataset~\cite{deng2009imagenet} for image classification.

As shown in Fig. \ref{fig:hcl}, to build the positive and negative pairs, we pair the restored images $\{A_{1},A_{2},A_{3}\}$ by HDN and their corresponding ground truth images $\{P_{1},P_{2},P_{3}\}$ (positive samples) in different resolutions as positive-oriented supervision to guide HDN to recover haze-free images. We take the hazy input images $\{N_{1},N_{2},N_{3}\}$ together with the restored images as negative pairs to enforce HDN to focus more on learning the complex haze components. The hierarchical contrastive loss is represented as:  
\begin{equation} 
\mathcal{L}_{\text {hcl}} = \sum_{i=1}^{3}\left(\sum_{j=1}^{3} \left\|\hat{A_{i}}-\hat{P_{j}}\right\|_{1}\right)\left(\sum_{k=1}^{3}\frac{1}{\left\|\hat{A_{i}}-\hat{N_{k}}\right\|_{1}}\right),
\end{equation}
where $\hat{N},\hat{A},\hat{P}$ denote the extracted features from the VGG-19 network and $\left\|.\right\|_{1}$ is $L_{1}$ norm to measure the similarity between two extracted features. Such a special design is expected to capture the relationship between recovered images in different resolutions. In the implementation, we extract the features from the 1st, 3rd, 5th, 9th, and 13th layers of the pre-trained VGG-19 and set the corresponding coefficients as $\frac{1}{32}$, $\frac{1}{16}$, $\frac{1}{8}$, $\frac{1}{4}$, and 1, respectively, which have been shown to be effective in image dehazing~\cite{liu2019griddehazenet,wu2021contrastive}. These features contain high-level semantic information and can effectively represent the global and local features of the image. In addition, when dealing with the cross-scale images, we interpolate images with different scales to the middle scale to calculate the loss. Our hierarchical contrastive loss aims to use the information from ground-truth images as an upper bound and treat hazy images as a lower bound to leverage the information from positive and negative images for image dehazing. Therefore, it is important to note that the reference embeddings in our loss are the generated images, not the ground-truth images.

\subsection{Loss Function}\label{sec:loss_function}
To train our proposed HDN, we design a loss function combining the Charbonnier loss ~\cite{charbonnier1994two} and the proposed hierarchical contrastive loss. We regard the Charbonnier loss as a pixel-wise loss, which is used between the recovered images and the ground truth images at each scale. The Charbonnier loss is defined as:
\begin{equation}
\mathcal{L}_{\text{char}}=\frac{1}{3}\sum_{k=1}^{3}\sqrt{\|A_{k}-P_{k}\|^{2}+\varepsilon^{2}},
\end{equation}
where $A_{k}$ and $P_{k}$ represent the dehazed image and ground-truth image respectively. $k$ is the index of the image scale level in the model. The constant $\varepsilon$ is empirically set to $10^{-3}$. With this constant, the network trained with Charbonnier loss can better handle outliers and improve performance over $\mathcal{L}_{2}$ loss function. In addition, the network can converge faster~\cite{charbonnier1994two}. The final loss function $\mathcal{L}_{\text{total}}$ to train our proposed HDN is determined as follows:
\begin{equation}
\mathcal{L}_{\text{total}}=\mathcal{L}_{\text {char}}+ \lambda \mathcal{L}_{\text {hcl}},
\end{equation}
where $\mathcal{L}_{\text {char}}$ is the Charbonnier loss, $\mathcal{L}_{\text {hcl}}$ is the proposed hierarchical contrastive loss. $\lambda$ is a hyper-parameter to balance these two loss terms, which is empirically set to $0.1$.

\section{Experiments}\label{sec:experiment}
In this part, we first explain the implementation details of the proposed method. Then, we show the image dehazing results of our approach and the comparison with the state-of-the-art methods. Finally, we conduct extensive ablation studies to verify the effectiveness of modules in the proposed method.

\subsection{Implementation Details}
\textbf{Datasets.} We evaluate the proposed method on synthetic and real-world datasets. RESIDE \cite{li2018benchmarking} is a large-scale synthetic dataset including indoor and outdoor scenes. This dataset contains five subsets, \ie Indoor Training Set (ITS), Outdoor Training Set (OTS), Synthetic Objective Testing Set (SOTS), Real World task-driven Testing Set (RTTS), and Hybrid Subjective Testing Set (HSTS). ITS contains $13,990$ hazy images and $1,399$ clear images. SOTS consists of $500$ indoor and outdoor paired images respectively. Following the previous works \cite{qin2020ffa,liu2019griddehazenet,dong2020multi}, we adopt ITS and OTS to train the proposed method respectively, and use SOTS and RTTS for performance evaluation. Furthermore, we test the proposed method on more challenging datasets, including HazeRD \cite{zhang2017hazerd} and DENSE-HAZE \cite{ancuti2019dense}, which are collected in real-world scenarios.

\textbf{Experimental Details.} In the experiment, we augment the training data with random rotations of $90$, $180$, and $270$. We randomly crop a $240 \times 240$ patch in the training stage. The batch size is set to $16$ and all weights of the models are initialized by the Xavier method. The learning rate is $2\times 10^{-4}$, which is steadily decreased to $1\times 10^{-6}$ by the cosine annealing strategy. We train all models for 400 epochs. And models are optimized by the Adam optimizer, where $\beta_{1}$ and $\beta_{2}$ are set to $0.9$ and $0.999$ respectively. In addition,  we adopt the Pytorch framework to perform all experiments on the NVIDIA Tesla V100 GPU.

\textbf{Comparison Methods.} We compare performance  of our method with seven state-of-the-art image dehazing methods: DCP \cite{he2010single}, DehazeNet \cite{cai2016dehazenet}, AOD-Net \cite{li2017aod}, GridDehazeNet \cite{liu2019griddehazenet},
FFA-Net \cite{qin2020ffa}, MSBDN \cite{dong2020multi}, and AECR-Net \cite{wu2021contrastive}.

\begin{table*}[t] 
	\begin{center}
	 \caption{Comparison results with the state-of-the-art image dehazing approaches on the benchmark datasets. The best and the second best performances are highlighted and underlined respectively. The proposed method achieves the best performance compared with previous state-of-the-arts.}
		\scalebox{0.7}{\begin{tabular}{c|cccccccc}
			 \toprule[1pt]
			 \multirow{2}{*}{Methods} & \multicolumn{2}{c}{ SOTS(indoor/outdoor) } & \multicolumn{2}{c}{ SOTS-average} & \multicolumn{2}{c}{ HazeRD} & \multicolumn{2}{c}{DENSE-HAZE}  \\
			
			 & PSNR $\uparrow$ & SSIM $\uparrow$  & PSNR $\uparrow$ & SSIM $\uparrow$  & PSNR $\uparrow$  & SSIM $\uparrow$ & PSNR $\uparrow$  & SSIM $\uparrow$ \\

			\hline DCP \cite{he2010single} & $15.16/13.85$ & $0.8546/0.5416$ & $13.85$ & $0.6516$ & $15.22$ & $0.7737$ & $10.06$ & $0.3856$ \\
			DehazeNet \cite{cai2016dehazenet} & $19.82/24.75$ & $0.8209/0.9269$ & $22.29$ & $0.8739$ & $15.63$ & $0.7517$ & $ 13.84$ & $0.4252$  \\
			AOD-Net \cite{li2017aod} & $20.51/24.14$ & $0.8162/0.9198$ & $22.33$ & $0.8680$ & $15.54$ & $0.7449$ & $13.14$ & $0.4144$ \\
			GridDehazeNet \cite{liu2019griddehazenet} & $32.16/30.86$ & $0.9836/0.9819$ & $31.51$ & $0.9828$ & $\underline{16.05}$ & $0.7932$ & $13.31$ & $0.3681$  \\
			FFA-Net \cite{qin2020ffa} & $36.39/\underline{33.57}$ & $0.9886/\underline{0.9840}$ & $\underline{34.98}$ & $0.9698$ & $14.96$ & $0.7654$ & $14.39$ & $0.4524$  \\
			MSBDN \cite{dong2020multi} & $-/-$ & $-/-$ & $33.79$ & $\underline{0.9840}$ & $15.78$ & $\underline{0.7982}$ & $15.37$ & $\underline{0.4858}$\\
			AECR-Net \cite{wu2021contrastive} & $\underline{37.17}/-$& $\underline{0.9901}/-$ & $-$ & $-$ & $-$ & $-$ & $\underline{15.80}$ & $0.4660$  \\
			\hline \textbf{HCD} & $\mathbf{38.31}$/$\mathbf{34.02}$ & $\mathbf{0.9954}/\mathbf{0.9936}$ & $\mathbf{36.16}$ & $\mathbf{0.9945}$ & $\mathbf{16.88}$ & $\mathbf{0.8088}$ & $\mathbf{16.41}$ & $\mathbf{0.5662}$ \\
			\bottomrule[1pt]
		\end{tabular}}
		\label{tab:results}
	\end{center}
\end{table*}

\subsection{Evaluation on Synthetic Datasets}
We evaluate our method on two synthetic dehazing datasets (\ie SOTS and HazeRD). 
The $2^{nd}$ and $3^{rd}$ columns in Table \ref{tab:results} show quantitative results on SOTS dataset in terms of PSNR and SSIM. As reported in the Table, DCP \cite{he2010single}, DehazeNet and AOD-Net present low values of PSNR and SSIM in both indoor and outdoor subsets, indicating that the dehazing results with low quality are produced. Compared with the previous methods, recent end-to-end methods (GridDehazNet \cite{liu2019griddehazenet}, FFA-Net \cite{qin2020ffa}, MSBDN \cite{dong2020multi}, AECR-Net \cite{wu2021contrastive}, and HCD) obtain better performance. Among them, the dehazing performance values of our proposed method rank first in both indoor and outdoor subsets of SOTS regarding PSNR and SSIM. For the indoor subset, our model surpasses the second best method (AECR-Net \cite{wu2021contrastive}) over $1.14$ dB, $0.0053$ on PSNR and SSIM respectively. For the outdoor subset, our model achieves $0.45$ dB and $0.0096$ improvement in terms of PSNR and SSIM. As for a more challenging dataset HazeRD, the evaluation results are shown in the $4^{th}$ column of Table \ref{tab:results}. We find that the performance of some learning-based methods is worse than that of the SOTS dataset. For example, FFA-Net \cite{qin2020ffa} obtains PSNR of $14.96$ dB and SSIM of $0.7654$, which is even worse than DCP \cite{he2010single}. This may be attributed to that, the number of samples in the HazeRD dataset is not sufficient enough to train the models well. Among all comparison methods, our HCD achieves the best dehazing performance in terms of PSNR and SSIM. The comparison reveals that our method achieves the highest performance for image dehazing.

\begin{table*}[t]
	\begin{center}
	 \caption{Quantitative comparison on SOTS-Indoor and DENSE-HAZE datasets. The best and second best performances are bold and underlined respectively. To ensure the fairness of comparison, we calculate MACs based on $256 \times 256$ color images.}
		\scalebox{0.9}{\begin{tabular}{|c|cc|cc|cc|}
			 \hline
			 \multirow{2}{*}{Methods} & \multicolumn{2}{c|}{ SOTS (indoor) } &  \multicolumn{2}{c|}{DENSE-HAZE} &\multicolumn{2}{c|}{Overhead} \\
			\cline{2-7}
			 & PSNR $\uparrow$ & SSIM $\uparrow$  & PSNR $\uparrow$  & SSIM $\uparrow$ & \#Param & MACs\\
			\hline 
			AECR-Net~\cite{wu2021contrastive} & $37.17$& $0.9901$ & $15.80$ & $0.4660$ &2.611 M & 52.20 G \\
			AECR-Net (large) & \underline{$37.96$} & \underline{$0.9941$} & \underline{$16.02$} & $0.5592$ &8.570 M & 105.41 G \\
			\hline \textbf{HCD (tiny)} & $37.62$ & $0.9940$ & $15.86$ & \underline{$0.5603$} & 2.580 M &51.04 G \\
			\textbf{HCD} & $\mathbf{38.31}$ & $\mathbf{0.9954}$ & $\mathbf{16.41}$ & $\mathbf{0.5662}$ & 5.580 M &104.03 G\\
		    \hline
		\end{tabular}}
		\label{tab:results_aecr-net}
	\end{center}
\end{table*}
To compare with the state-of-the-art AECR-Net, we conduct experiments from the following two aspects: 1) Increasing the complexity of AECR-Net (called AECR-Net (large)) to make it close to the proposed HCD for performance comparison; 2) Reducing the complexity of our proposed model (named HCD (tiny)), and then compares the performance with AECR-Net. The results are shown in Table. \ref{tab:results_aecr-net} demonstrate that the proposed HCD achieves better performance and complexity trade-off (\ie 36.16 dB, 5.58 M, 104.03 G) than the state-of-the-art AECR-Net.

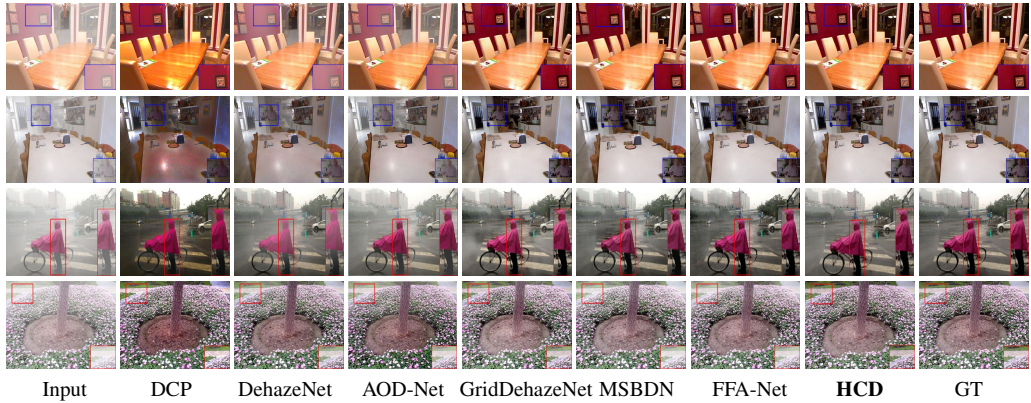
\begin{figure*}[t]
\begin{center}
 \begin{overpic}[width=\textwidth]{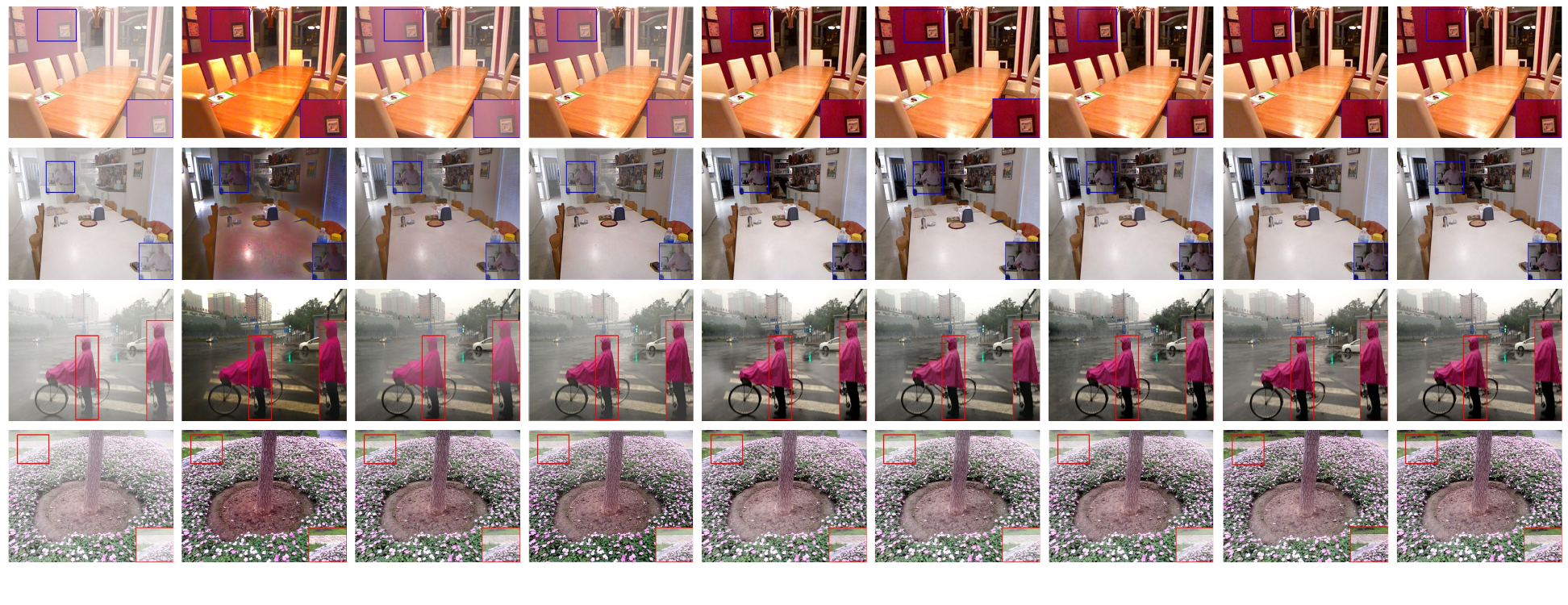}
    \put(4,0){\scriptsize Input}
   \put(14.5,0){\scriptsize DCP}
   \put(23,0){\scriptsize DehazeNet}
   \put(35,0){\scriptsize AOD-Net}
   \put(44.5,0){\scriptsize GridDehazeNet}
   \put(58,0){\scriptsize MSBDN}
   \put(69,0){\scriptsize FFA-Net}
   \put(81,0){\scriptsize \textbf{HCD}}
   \put(92.5,0){\scriptsize GT}
 \end{overpic}
 	\caption{Visual comparison with the state-of-the-art image dehazing methods on the SOTS dataset. The top two rows are images from the indoor subset, and the bottom two rows are images from the outdoor subset. \textbf{Zoom in for details}.}
	\label{fig:in_out_results}
\end{center}
\end{figure*}
We also demonstrate visual comparisons of the dehazed results from different methods. 
As shown in Fig. \ref{fig:in_out_results}, we present the haze removal results for all comparison methods in indoor and outdoor hazy images. In Fig. \ref{fig:in_out_results}, the top two rows are images from the indoor testing subset, and the bottom two rows correspond to the outdoor testing subset. From Fig. \ref{fig:in_out_results}, we find that DCP \cite{he2010single} and DehazeNet \cite{cai2016dehazenet} suffer from the color distortion problem so their dehazed results seem unrealistic. The recovered images of AOD-Net \cite{li2017aod} are darker than the ground truth images in some cases (\eg the dining table in the second column of Fig. \ref{fig:in_out_results}). Though GridDehazeNet \cite{liu2019griddehazenet}, MSBDN \cite{dong2020multi}, and FFA-Net \cite{qin2020ffa} perform well, the artifacts of incomplete haze removal still exist in the restored images, \eg the flower in the images of Fig. \ref{fig:in_out_results}. Compared with these methods, our method produces images with rich details and color information, and there are rare artifacts in the dehazed images. Overall, our method achieves the best performance among the comparative methods from both quantitative and qualitative aspects.

\begin{figure*}[t]
\begin{center}
	 \begin{overpic}[width=\textwidth]{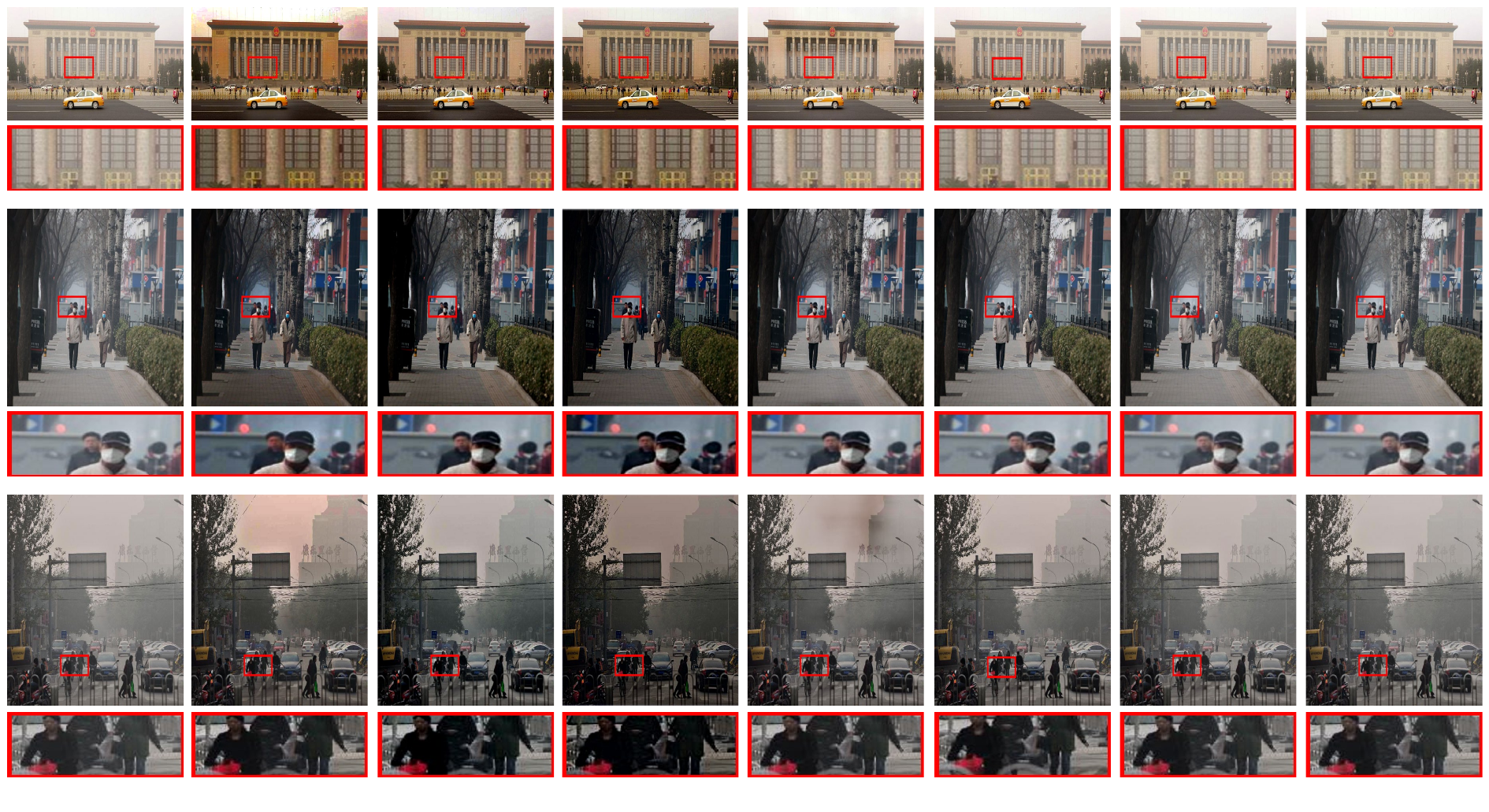} 
    \put(4,0){\scriptsize Input}
   \put(16.5,0){\scriptsize DCP}
   \put(27,0){\scriptsize DehazeNet}
   \put(40,0){\scriptsize AOD-Net}
   \put(50.5,0){\scriptsize GridDehazeNet}
   \put(65.5,0){\scriptsize MSBDN}
   \put(78,0){\scriptsize FFA-Net}
   \put(92,0){\scriptsize \textbf{HCD}}
 \end{overpic}
	\caption{Visual comparison on real-world hazy images from the RTTS dataset. \textbf{Zoom in for details}.}
	\label{fig:rtts_results}
	\end{center}
\end{figure*}
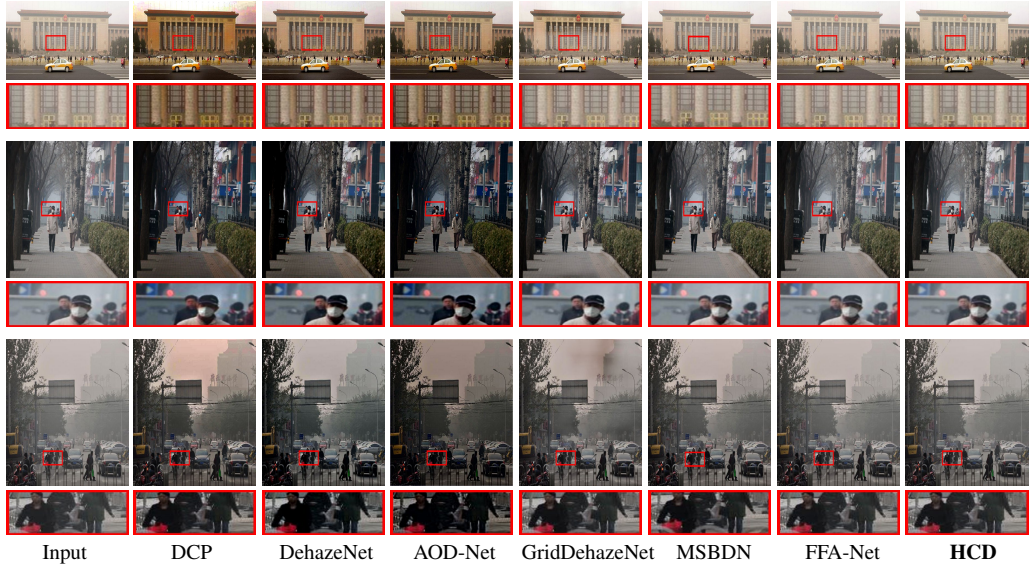
\subsection{Evaluation on Real World Datasets}
We further evaluate the proposed method quantitatively and qualitatively on real-world images. Specifically, we conduct quantitative experiments on the DENSE-HAZE dataset, in which hazy images are captured in natural scenes. As shown in Table \ref{tab:results}, our HCD achieves the best performance in terms of PSNR and SSIM. Our HCD significantly outperforms the second best method. To be specific, compared with the state-of-the-art AECR-Net \cite{wu2021contrastive}, our HCD achieves an advance of $0.61$ dB in terms of PSNR and surpasses MSBDN \cite{dong2020multi} by $0.08$ with regard to SSIM. 

\begin{table}[t]
	\begin{center}
	\caption{Ablation study of individual components. The values are average dehazed results of different variants on the indoor subset of the SOTS dataset. The best and second best performances are bold and underlined respectively.}
        \begin{tabular}{c|ccc|cc|c}
			 \toprule[1pt]
			 \multirow{2}{*}{Models} &  \multicolumn{3}{c|}{Components}  &\multicolumn{2}{c|}{SOTS-indoor} & \multirow{2}{*}{Params.} \\
			
			  & DCN & HFB & HCL &  PSNR $\uparrow$ & SSIM $\uparrow$\\
			\hline baseline & \XSolidBrush & \XSolidBrush & \XSolidBrush & $34.18$ &  $0.9926$ & $\mathbf{2.34}$ M\\
			baseline+DCN &  \CheckmarkBold  & \XSolidBrush & \XSolidBrush & $34.79$ & $0.9921$ & \underline{$4.04$} M \\
			baseline+DCN+HFB &  \CheckmarkBold  & \CheckmarkBold & \XSolidBrush & \underline{$36.69$} & \underline{$0.9931$} & $5.58$ M\\
			\hline \textbf{full model (HCD)} & \CheckmarkBold  & \CheckmarkBold  & \CheckmarkBold  & $\mathbf{38.31}$ & $\mathbf{0.9954}$ & $5.58$ M \\
			\bottomrule[1pt]
		\end{tabular}
		\label{tab:ablation}
	\end{center}
\end{table}

As for qualitative comparison, we compare the visual results of different methods on real-world hazy images from the RTTS dataset. As illustrated in Fig. \ref{fig:rtts_results}, prior-based method DCP \cite{he2010single} suffers from the color distortion problem (\eg the sky and street in the images of the second column in Fig. \ref{fig:rtts_results}). DehazeNet \cite{cai2016dehazenet} and AOD-Net \cite{li2017aod} cause color distortion in some scenery, and GridDehazeNet \cite{liu2019griddehazenet} encounters dark-area artifacts (\eg see the third image in the fifth column of Fig. \ref{fig:rtts_results}). Compared to the above methods, MSBAN \cite{dong2020multi} and FFA-Net \cite{qin2020ffa} produce better dehazing results. However, they still have the incomplete haze removal problem and produce some remaining haze artifacts in the restored images. In contrast, our model can work well and produce more visually plausible dehazing results in real-world scenery.

\subsection{Ablation Study}
We conduct extensive ablation studies to validate each component of our HCD. For a fair comparison, all models are trained on ITS with the same settings and are tested on the indoor subsets of the SOTS dataset.

\textbf{Individual components.} To better verify each component of our HCD, we conduct an ablation study by considering three factors: the first is the deformable convolution DCN in the hierarchical feature extractor. The second is the hierarchical fusion block (HFB) in the hierarchical interaction module and the last one is the hierarchical contrastive loss (HCL). Regarding these factors, we design the following variants to realize the ablation analysis: 1) baseline: it is a baseline network, which does not adopt any of the above components (\ie the deformable convolution DCN in the hierarchical feature extractor, HFB, and HCL). It means the network cannot benefit from the deformable convolution in the stage of feature extraction, the hierarchical feature fusion by HFB, and the contrastive learning.  
2) baseline+DCN: Add the deformable convolution into the hierarchical feature extractor of the baseline network. It means the network can extract more abundant features for image dehazing. 3) baseline+DCN+HFB: This model adds the HFB based on baseline+DCN. 4) HCD: it is our complete model, which is optimized by the proposed HCL and Charbonnier loss. HCL allows the network to use both negative and positive samples for training.  In addition, baseline, baseline+DCN, and baseline+DCN+HFB are trained only by the Charbonnier loss. The detailed configuration of these models can be found in Table~\ref{tab:ablation}.

\begin{table}[t]
	\begin{center}
	\caption{Ablation study of our HCL. $\mathcal{L}_{\text{char}}$ indicates using pixel-wise loss in a multi-scale manner, and $\mathcal{L}_{\text{hchar}}$ represents using pixel-wise loss in a hierarchical structure. $\mathcal{L}_{\text{hcl}}$ refers to the proposed contrastive loss. The best and second best performances are bold and underlined respectively.}
    \begin{tabular}{c|ccc|cc}
			 \toprule[1pt]
			 \multirow{2}{*}{Models} &  \multicolumn{3}{c|}{Components}  &\multicolumn{2}{c}{SOTS-indoor}  \\
			& $\mathcal{L}_{\text{char}}$ & $\mathcal{L}_{\text {hchar}}$ & $\mathcal{L}_{\text {hcl}}$ &  PSNR $\uparrow$ & SSIM $\uparrow$\\
			\hline 
			Model1 &  \CheckmarkBold  & \XSolidBrush & \XSolidBrush & $36.69$ & $0.9931$ \\
			Model2 &  \XSolidBrush  & \CheckmarkBold & \XSolidBrush & $\underline {37.01}$ & $\underline {0.9932}$  \\
			\hline \textbf{HCD} & \CheckmarkBold  & \XSolidBrush & \CheckmarkBold  & $\mathbf{38.31}$ & $\mathbf{0.9954}$  \\
			\bottomrule[1pt]
		\end{tabular}
		\label{tab:ablation2}
		\vspace{-0.3in}
	\end{center}
\end{table}	
The values of PSNR and SSIM are represented in Table \ref{tab:ablation}. The performance of our complete model HCD shows great superiority over its incomplete counterparts, including baseline+DCN+HFB (removing HCL component), baseline+DCN (removing HCL and HFB), and baseline (removing HCL, HFB, and $\operatorname{DCN}$). Comparing baseline and baseline+DCN, the results show that embedding $\operatorname{DCN}$ in the hierarchical feature extractor leads to better performance. Moreover, for baseline+DCN+HFB, it surpasses baseline+DCN by $1.89$ dB, $0.001$ in terms of PSNR and SSIM. The result verifies that designing a reasonable fusion module in the model is important. Finally, HCD gains an evident improvement regarding baseline+DCN+HFB, proving that the proposed hierarchical contrastive learning effectively guides the network to enhance feature representation for image dehazing. In conclusion, each component of this work contributes to improving the dehazing performance.

\textbf{Ablation study about our HCL.} We design a pixel-wise loss in a hierarchical manner like contrastive loss and train the proposed network with the designed hierarchical pixel-wise loss. The results of the derived Model2 are shown in Table \ref{tab:ablation2}. Model2 is trained with only the hierarchical pixel-wise loss, and there is only a slight performance improvement compared to Model1 (trained using pixel-wise loss in a multi-scale manner). The comparison results show that the performance improvement of dehazing can indeed be attributed to using the proposed contrastive loss. 

\begin{table}[t]
	\begin{center}
	 \caption{Comparison results of the state-of-the-art image deraining approaches on the benchmark datasets. The best and second best performances are bold and underlined respectively. }
          \begin{tabular}{c|cccc}
			 \toprule[1pt]
			 \multirow{2}{*}{Methods} &  \multicolumn{2}{c}{Test100} & \multicolumn{2}{c}{Rain100L}  \\
			
			  & PSNR $\uparrow$ & SSIM $\uparrow$  & PSNR $\uparrow$  & SSIM $\uparrow$ \\

			\hline DerainNet \cite{fu2017clearing} & $22.77$ & $0.810$ & $27.03$ & $0.884$  \\
			SEMI  \cite{wei2019semi} &  $22.35$ & $0.788$ & $25.03$ & $0.842$  \\
			DIDMDN \cite{zhang2018density} &  $22.56$ & $0.818$ & $25.23$ & $0.741$ \\
			UMRL \cite{yasarla2019uncertainty} & $24.41$ & $0.829$ & $29.18$ & $0.923$  \\
			RESCAN \cite{li2018recurrent} &  $25.00$ & $0.835$ & $29.80$ & $0.881$   \\
			PReNet \cite{ren2019progressive} &  $24.81$ & $0.851$ & $\underline{32.44}$ & $\underline{0.950}$ \\
			MSPFN \cite{jiang2020multi} &  $\underline{27.50}$ & $\underline{0.876}$ & $32.40$ & $0.933$   \\
\hline \textbf{HCD} & $\mathbf{29.43}$ & $\mathbf{0.892}$ & $\mathbf{35.01}$ & $\mathbf{0.956}$   \\
			\bottomrule[1pt]
		\end{tabular}
		\label{tab:deraining}
	\end{center}
\end{table}
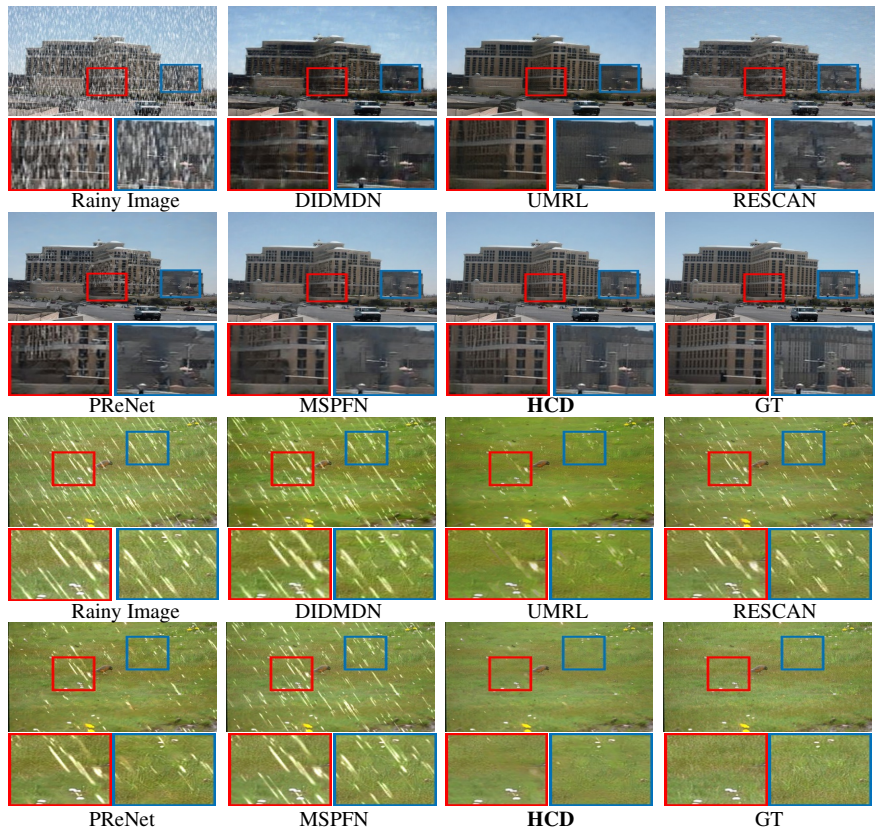
\begin{figure*}[t]
\begin{center}
\begin{overpic}[width=0.85\textwidth]{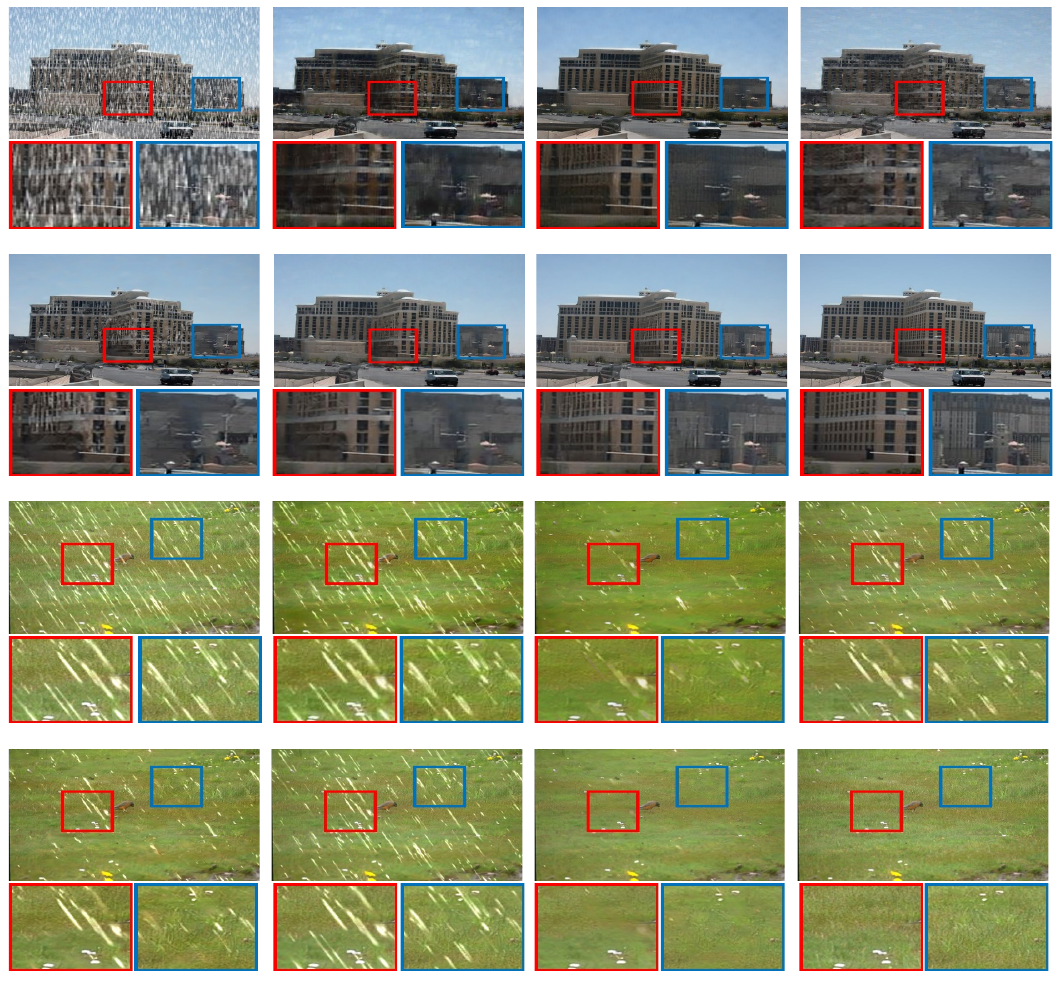} 
   \put(8,71.7){\scriptsize Rainy Image}
   \put(33.5,71.7){\scriptsize DIDMDN}
    \put(60,71.7){\scriptsize UMRL}
    \put(83.5,71.7){\scriptsize RESCAN}
    \put(10,48.2){\scriptsize PReNet}
    \put(34,48.2){\scriptsize MSPFN}
    \put(60,48.2){\scriptsize \textbf{HCD}}
      \put(86,48.2){\scriptsize GT}
  \put(8,24.7){\scriptsize Rainy Image}
  \put(33.5,24.7){\scriptsize DIDMDN}
  \put(60,24.7){\scriptsize UMRL}
  \put(83.5,24.7){\scriptsize RESCAN}
  \put(10,1){\scriptsize PReNet}
  \put(34,1){\scriptsize MSPFN}
  \put(60,1){\scriptsize \textbf{HCD}}
  \put(86,1){\scriptsize GT}
 \end{overpic}
\caption{Visual comparison for the image deraining task on the Test100 and Rain100L datasets~\cite{yang2017deep}. The top two rows of images are from the Test100 dataset, and the bottom two rows are images from the Rain100L dataset. The images generated by our HCD have almost no rainy drops, which are clear and more similar to the GT images.
}
	\label{fig:deraining_results}
	\end{center}
	\vspace{-0.2in}
\end{figure*}

\begin{table}[t]
	\begin{center}
	 \caption{Comparison results for the nighttime dehazing task. The best and second best performances are bold and underlined respectively.} 
		\begin{tabular}{c|ccc}
			 \toprule[1pt]
			 \multirow{2}{*}{Methods} &  \multicolumn{3}{c}{NHR} \\
			
			  & PSNR $\uparrow$ & SSIM $\uparrow$  & LPIPS $\downarrow$ \\
            
            \hline NDIM~\cite{zhang2014nighttime}  & $14.31$ & $0.5256$ & $-$   \\
             GS~\cite{li2015nighttime}  & $17.32$ & $0.6285$ & $-$    \\
             MRPF~\cite{zhang2017fast}  & $16.95$ & $0.6674$ &  $-$   \\
			MPR \cite{zhang2017fast} & $19.93$ & $0.7772$ &  $0.3072$  \\
			OSFD  \cite{zhang2020nighttime} &  \underline{$21.32$} & \underline{$0.8035$} & \underline{$0.2910$} \\
		\hline \textbf{HCD (Ours)} & $\mathbf{23.43}$ & $\mathbf{0.9533}$  & $\mathbf{0.0729}$ \\
			\bottomrule[1pt]
		\end{tabular}
		\label{tab:neighttime}
	\end{center}
\end{table}
\subsection{Generalization on Other Tasks} %
\textbf{Image Deraining.} To explore the generalization of the proposed method, we consider adopting our model for the task of image deraining, which is similar to image dehazing. Following the previous work \cite{jiang2020multi}, we use synthetic paired rainy images to train our model and test the performance on both Test100 \cite{zhang2019image} and Rain100L \cite{yang2017deep} testing sets. 
The training strategy is the same as the image dehazing task. We adopt PSNR and SSIM metrics to evaluate the deraining performance. In this task, PSNR and SSIM are calculated on the Y channel in the YCbCr color space like \cite{jiang2020multi}. We choose seven state-of-the-art image deraining methods for comparison: 1) deraining network (DerainNet) \cite{fu2017clearing}, 2) semi-supervised deraining method (SEMI) \cite{wei2019semi}, 3) density aware multi-stream densely connected convolutional neural network (DIDMDN) \cite{zhang2018density}, 4) uncertainty guided multi-scale residual learning network (UMRL) \cite{yasarla2019uncertainty}, 5) recurrent squeeze-and-excitation context aggregation network (RESCAN) \cite{li2018recurrent}, 6) progressive
recurrent network (PReNet) \cite{ren2019progressive}, and 7) multi-scale progressive fusion network (MSPFN) \cite{jiang2020multi}.

The quantitative results on Test100 and Rain100L are reported in Table \ref{tab:deraining}. Our HCD achieves remarkable performance on both Test100 and Rain100L datasets. Especially, on the Rain100L dataset, our HCD surpasses MSPFN by $2.61$ dB and $0.023$ in terms of PSNR and SSIM. We also show the visual comparison in Fig. \ref{fig:deraining_results}. We note that the state-of-the-art deraining methods (\eg PReNet and MSPFN) do not remove the rain streaks well (see patches of restored images in Fig. \ref{fig:deraining_results}). In contrast, our HCD can remove the rain well and recover high-quality images with truthful details compared to other methods. The comparison results demonstrate our HCD generalizes well on image deraining, though our HCD is specifically designed for image dehazing.

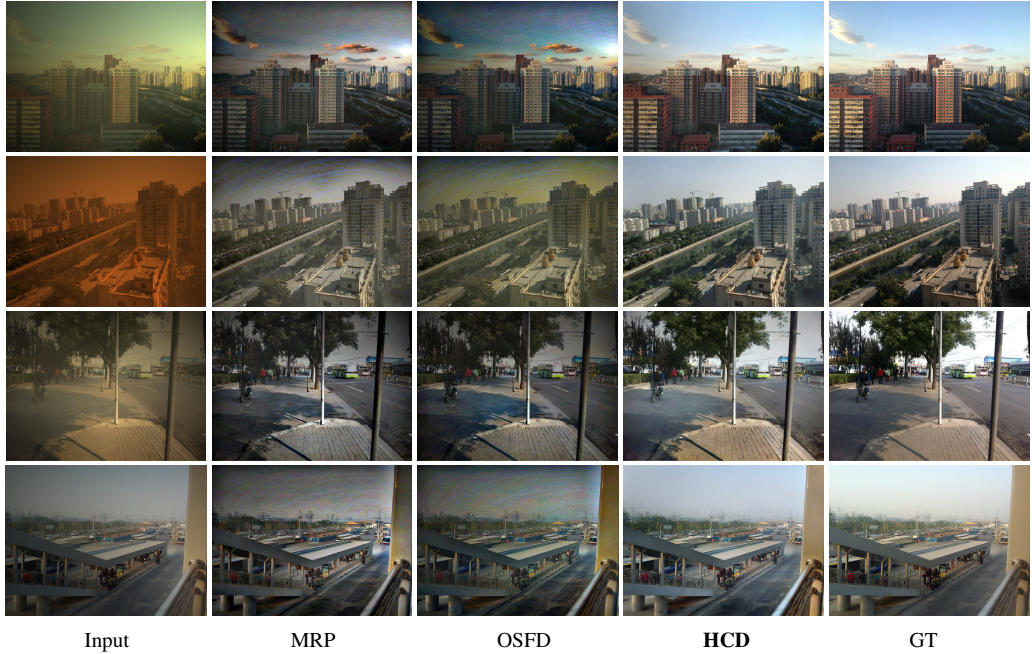
\begin{figure*}[t]
\begin{center}
\begin{overpic}[width=\textwidth]{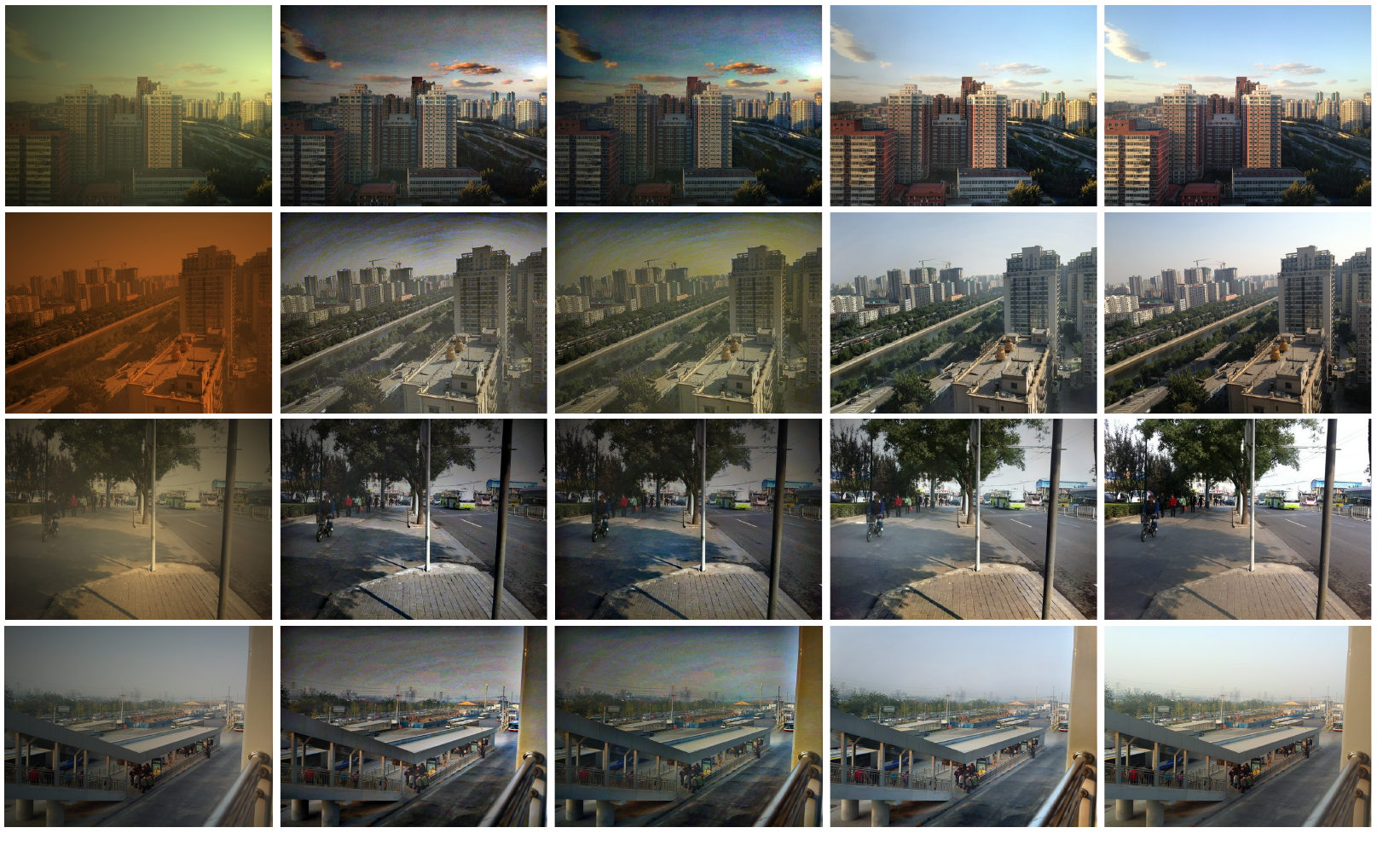} 
    \put(8,0){\scriptsize Input}
   \put(28,0){\scriptsize MRP}
   \put(48,0){\scriptsize OSFD}
   \put(68,0){\scriptsize \textbf{HCD}}
   \put(88,0){\scriptsize GT}
 \end{overpic}
\caption{Visual examples for the nighttime dehazing task on the NHR dataset~\cite{zhang2020nighttime}. Our HCD can successfully remove the haze in low-light conditions and does not introduce artifacts in the recovered images.}
	\label{fig:nighttime_dehazing_results}
	\end{center}
\end{figure*}
\textbf{Nighttime Dehazing.} We further investigate the potential of our proposed method on the nighttime dehazing task. To be specific, we adopt the NHR dataset \cite{zhang2020nighttime} to train and test our proposed method. NHR contains $17,940$ pairs of images. We choose $1,794$ pairs of images for evaluation and other samples for training, following the previous work \cite{zhang2020nighttime}. For comparison, we select five representative nighttime image dehazing methods, including
NDIM~\cite{zhang2014nighttime}, GS~\cite{li2015nighttime}, MRPF~\cite{zhang2017fast}, MPR~\cite{zhang2017fast}, and OSFD~\cite{zhang2020nighttime}.

The quantitative results are reported in Table. \ref{tab:neighttime}. We use both pixel-wise (PSNR and SSIM) and perceptual (LPIPS) metrics to evaluate the performance. As listed in Table \ref{tab:neighttime}, HCD achieves the best performance in terms of PSNR, SSIM, and LPIPS, respectively. 
Especially, the advance of HCD is $2.11$ dB and $0.1498$ in terms of PSNR and SSIM compared to OSFD \cite{zhang2020nighttime} that is the best method among the comparison methods. 
The perceptual metric LPIPS further demonstrates the superiority of our HCD.
Fig. \ref{fig:nighttime_dehazing_results} provides a visual comparison. Compared with state-of-the-art methods (MPR \cite{zhang2017fast} and OSFD \cite{zhang2020nighttime}), the proposed HCD can effectively remove the haze and recover images with better brightness at the same time, and it introduces fewer artifacts. Quantitative and qualitative results indicate that the proposed HCD has a strong capability to process hazy images under low-light conditions.

\subsection{Discussion about the proposed HCL}

To further test the universality of the proposed hierarchical loss, we add the HCL into the existing state-of-the-art methods including GridDehazeNet~\cite{liu2019griddehazenet}, FFA-Net~\cite{qin2020ffa}, MSBDN~\cite{dong2020multi},  and AECR-Net~\cite{wu2021contrastive} for performance evaluation. As shown in Table.~\ref{tab:results_add}, our proposed loss can indeed improve the performance of SOTS methods. For example, GridDehazeNet~\cite{liu2019griddehazenet} achieves  higher PSNR and SSIM with gains of $1.08$ and $0.0051$, respectively. This universal experimental validation shows that the proposed loss does not rely on a particular network and it can train the dehazing network effectively.

\begin{table}[t]
	\begin{center}
	 \caption{Quantitative results when applying the proposed contrastive loss into the existing state-of-the-art dehazing methods.}
		\scalebox{0.9}{\begin{tabular}{|c|cccc|}
		\hline
			 \multirow{2}{*}{Methods} & \multicolumn{4}{c|}{ SOTS (indoor) }  \\
			\cline{2-5}
			 & \multicolumn{2}{c|}{PSNR}  & \multicolumn{2}{c|}{SSIM}   \\

			\hline 
			GridDehazeNet~\cite{liu2019griddehazenet} & \multicolumn{2}{c|}{33.24 ( \color{red}{1.08 $\uparrow$})}& \multicolumn{2}{c|}{0.9887 ( \color{red}{0.0051 $\uparrow$})}  \\
			FFA-Net~\cite{qin2020ffa} & \multicolumn{2}{c|}{36.96 ( \color{red}{0.57 $\uparrow$})}& \multicolumn{2}{c|}{0.9908 (\color{red}{ 0.0022 $\uparrow$})}  \\
			MSBDN~\cite{dong2020multi} & \multicolumn{2}{c|}{34.66 ( \color{red}{0.87 $\uparrow$})}& \multicolumn{2}{c|}{0.9889 ( \color{red}{0.0049 $\uparrow$})}  \\
			AECR-Net~\cite{wu2021contrastive} & \multicolumn{2}{c|}{37.83 ( \color{red}{0.66 $\uparrow$})}& \multicolumn{2}{c|}{0.9912 ( \color{red}{0.0011 $\uparrow$})}  \\
            \hline
		\end{tabular}}
		\label{tab:results_add}
		\vspace{-0.1in}
	\end{center}
\end{table}

\begin{figure}[t]
\begin{center}
  \begin{overpic}[width=\textwidth]{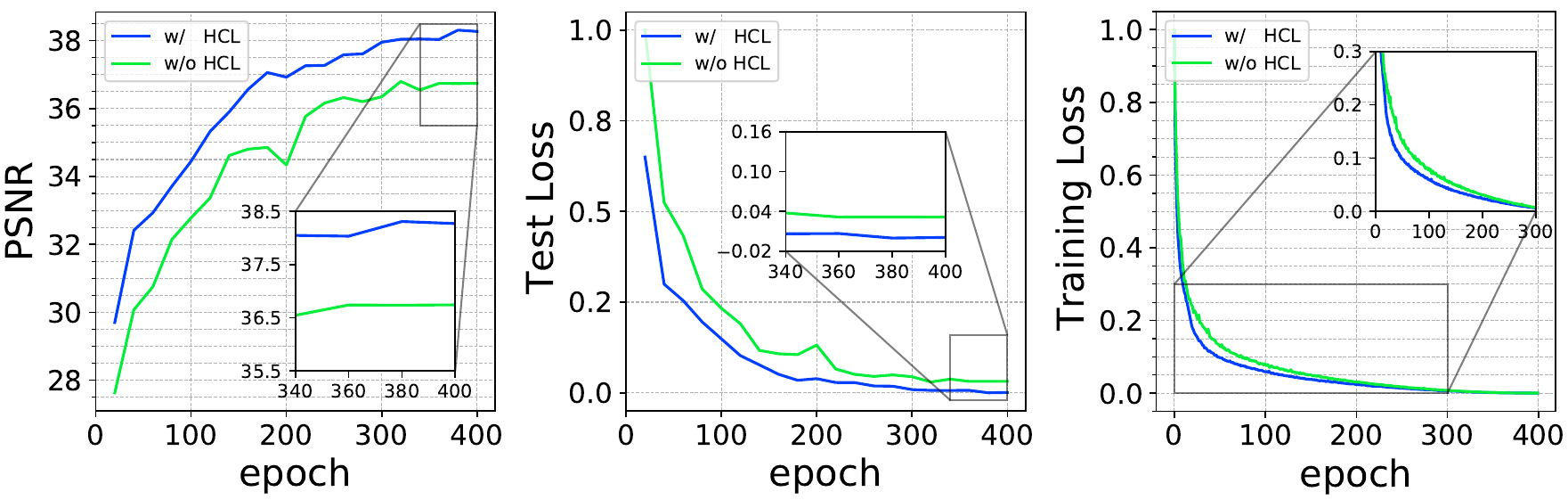} 
    \put(17,-2){(a)}
  \put(51,-2){(b)}
  \put(85,-2){(c)}
 \end{overpic}
 	\caption{Quantitative comparison between the proposed models trained with and without HCL. (a) Results in the SOTS-indoor testing set. (b) Test loss curve. (c) Training loss curve. The model trained with HCL achieves better performance.}
	\label{fig:loss_psnr}
	\end{center}
\end{figure}

To demonstrate the effectiveness of our proposed HCL, we conduct an analysis to assess whether it enhances the model's performance. To validate this, we perform experiments by removing HCL (as shown in Fig.~\ref{fig:loss_psnr} (a), (b) and (c)). The results consistently show that while HCL may not significantly accelerate the convergence speed during training, the model trained with HCL consistently achieves higher PSNR values on the testing set compared to the model trained without it. This substantial improvement in PSNR values suggests that our proposed HCL positively impacts the model's performance, enhancing its ability to produce superior image restorations.

\section{Conclusion}\label{sec:conclusion}
In this work, we propose a novel hierarchical contrastive dehazing (HCD) method, which consists of a hierarchical dehazing network (HDN) and a hierarchical contrastive loss (HCL). In HDN, we propose a hierarchical interaction module, which effectively fuses the hierarchical features so that the learned features can better facilitate haze removal. To further remove the haze component from the input image, our special hierarchical HDN performs hierarchical contrastive learning by constructing the positive and negative pairs in a hierarchical manner. Extensive experimental results on synthetic benchmarks and real-world images have shown the great superiority of our HCD over state-of-the-art methods.

However, similar to other image dehazing methods, our method still has some limitations under real-world application. Accordingly, there are multiple directions to explore in the future. First, due to our method relying on paired data, we will explore the unsupervised learning strategy, which can help our method easily deal with the image dehazing problem on real-world scenery. Second, we will extend our method to many other relevant image restoration tasks, such as image desnowing, image denoising, image deblurring, and low-light image enhancement.

\section*{Acknowledgement}
This work was supported in part by the National Natural Science Foundation of China (Grant No. 62372223), and in part by the Fundamental Research Funds for the Central Universities (No. XJ2023000701).

\bibliographystyle{elsarticle-num}

\bibliography{references}

\begin{thebibliography}{10}
\expandafter\ifx\csname url\endcsname\relax
  \def\url#1{\texttt{#1}}\fi
\expandafter\ifx\csname urlprefix\endcsname\relax\def\urlprefix{URL }\fi
\expandafter\ifx\csname href\endcsname\relax
  \def\href#1#2{#2} \def\path#1{#1}\fi

\bibitem{ali2023boundary}
U.~Ali, J.~Choi, K.~Min, Y.-K. Choi, M.~T. Mahmood, Boundary-constrained robust regularization for single image dehazing, Pattern Recognition 140 (2023) 109522.

\bibitem{tan2008visibility}
R.~T. Tan, Visibility in bad weather from a single image, in: Proceedings of IEEE Conference on Computer Vision and Pattern Recognition, 2008, pp. 1--8.

\bibitem{li2017aod}
B.~Li, X.~Peng, Z.~Wang, J.~Xu, D.~Feng, Aod-net: All-in-one dehazing network, in: Proceedings of IEEE International Conference on Computer Vision, 2017, pp. 4770--4778.

\bibitem{cai2016dehazenet}
B.~Cai, X.~Xu, K.~Jia, C.~Qing, D.~Tao, Dehazenet: An end-to-end system for single image haze removal, IEEE Transactions on Image Processing 25~(11) (2016) 5187--5198.

\bibitem{li2022dual}
Z.~Li, C.~Zheng, H.~Shu, S.~Wu, Dual-scale single image dehazing via neural augmentation, IEEE Transactions on Image Processing 31 (2022) 6213--6223.

\bibitem{lin2022msaff}
C.~Lin, X.~Rong, X.~Yu, Msaff-net: Multi-scale attention feature fusion networks for single image dehazing and beyond, IEEE Transactions on Multimedia (2022).

\bibitem{jiang2023deep}
N.~Jiang, K.~Hu, T.~Zhang, W.~Chen, Y.~Xu, T.~Zhao, Deep hybrid model for single image dehazing and detail refinement, Pattern Recognition 136 (2023) 109227.

\bibitem{liu2023local}
Y.~Liu, X.~Hou, Local multi-scale feature aggregation network for real-time image dehazing, Pattern Recognition (2023) 109599.

\bibitem{sun2023multi}
H.~Sun, B.~Li, Z.~Dan, W.~Hu, B.~Du, W.~Yang, J.~Wan, Multi-level feature interaction and efficient non-local information enhanced channel attention for image dehazing, Neural Networks 163 (2023) 10--27.

\bibitem{hariharan2015hypercolumns}
B.~Hariharan, P.~Arbel{\'a}ez, R.~Girshick, J.~Malik, Hypercolumns for object segmentation and fine-grained localization, in: Proceedings of IEEE Conference on Computer Vision and Pattern Recognition, 2015, pp. 447--456.

\bibitem{qin2020ffa}
X.~Qin, Z.~Wang, Y.~Bai, X.~Xie, H.~Jia, Ffa-net: Feature fusion attention network for single image dehazing, in: Proceedings of AAAI Conference on Artificial Intelligence, 2020, pp. 11908--11915.

\bibitem{qu2019enhanced}
Y.~Qu, Y.~Chen, J.~Huang, Y.~Xie, Enhanced pix2pix dehazing network, in: Proceedings of IEEE Conference on Computer Vision and Pattern Recognition, 2019, pp. 8160--8168.

\bibitem{zhang2020multi}
X.~Zhang, T.~Wang, W.~Luo, P.~Huang, Multi-level fusion and attention-guided cnn for image dehazing, IEEE Transactions on Circuits and Systems for Video Technology 31~(11) (2020) 4162--4173.

\bibitem{wu2021contrastive}
H.~Wu, Y.~Qu, S.~Lin, J.~Zhou, R.~Qiao, Z.~Zhang, Y.~Xie, L.~Ma, Contrastive learning for compact single image dehazing, in: Proceedings of IEEE Conference on Computer Vision and Pattern Recognition, 2021, pp. 10551--10560.

\bibitem{li2018benchmarking}
B.~Li, W.~Ren, D.~Fu, D.~Tao, D.~Feng, W.~Zeng, Z.~Wang, Benchmarking single-image dehazing and beyond, IEEE Transactions on Image Processing 28~(1) (2018) 492--505.

\bibitem{zhang2017hazerd}
Y.~Zhang, L.~Ding, G.~Sharma, Hazerd: an outdoor scene dataset and benchmark for single image dehazing, in: Proceedings of IEEE International Conference on Image Processing, 2017, pp. 3205--3209.

\bibitem{liu2019griddehazenet}
X.~Liu, Y.~Ma, Z.~Shi, J.~Chen, Griddehazenet: Attention-based multi-scale network for image dehazing, in: Proceedings of IEEE Conference on Computer Vision and Pattern Recognition, 2019, pp. 7314--7323.

\bibitem{mutimbu2018factor}
L.~Mutimbu, A.~Robles-Kelly, A factor graph evidence combining approach to image defogging, Pattern Recognition 82 (2018) 56--67.

\bibitem{he2010single}
K.~He, J.~Sun, X.~Tang, Single image haze removal using dark channel prior, IEEE Transactions on Pattern Analysis and Machine Intelligence 33~(12) (2010) 2341--2353.

\bibitem{fattal2014dehazing}
R.~Fattal, Dehazing using color-lines, ACM Transactions on Graphics 34~(1) (2014) 1--14.

\bibitem{zhu2015fast}
Q.~Zhu, J.~Mai, L.~Shao, A fast single image haze removal algorithm using color attenuation prior, IEEE Transactions on Image Processing 24~(11) (2015) 3522--3533.

\bibitem{yuan2021confidence}
F.~Yuan, Y.~Zhou, X.~Xia, X.~Qian, J.~Huang, A confidence prior for image dehazing, Pattern Recognition 119 (2021) 108076.

\bibitem{gui2022comprehensive}
J.~Gui, X.~Cong, Y.~Cao, W.~Ren, J.~Zhang, J.~Zhang, J.~Cao, D.~Tao, A comprehensive survey and taxonomy on single image dehazing based on deep learning, ACM Computing Surveys (2022).

\bibitem{yin2020novel}
S.~Yin, Y.~Wang, Y.-H. Yang, A novel image-dehazing network with a parallel attention block, Pattern Recognition 102 (2020) 107255.

\bibitem{chen2021psd}
Z.~Chen, Y.~Wang, Y.~Yang, D.~Liu, Psd: Principled synthetic-to-real dehazing guided by physical priors, in: Proceedings of the IEEE Conference on Computer Vision and Pattern Recognition, 2021, pp. 7180--7189.

\bibitem{zhang2021single}
K.~Zhang, Y.~Li, Single image dehazing via semi-supervised domain translation and architecture search, IEEE Signal Processing Letters 28 (2021) 2127--2131.

\bibitem{zhu2017unpaired}
J.-Y. Zhu, T.~Park, P.~Isola, A.~A. Efros, Unpaired image-to-image translation using cycle-consistent adversarial networks, in: Proceedings of the IEEE International Conference on Computer Vision, 2017, pp. 2223--2232.

\bibitem{engin2018cycle}
D.~Engin, A.~Gen{\c{c}}, H.~Kemal~Ekenel, Cycle-dehaze: Enhanced cyclegan for single image dehazing, in: Proceedings of the IEEE Conference on Computer Vision and Pattern Recognition Workshops, 2018, pp. 825--833.

\bibitem{dudhane2019cdnet}
A.~Dudhane, S.~Murala, Cdnet: Single image de-hazing using unpaired adversarial training, in: Proceedings of the IEEE Winter Conference on Applications of Computer Vision, 2019, pp. 1147--1155.

\bibitem{park2020contrastive}
T.~Park, A.~A. Efros, R.~Zhang, J.-Y. Zhu, Contrastive learning for unpaired image-to-image translation, in: Proceedings of European Conference on Computer Vision, 2020, pp. 319--345.

\bibitem{zhang2021blind}
J.~Zhang, S.~Lu, F.~Zhan, Y.~Yu, Blind image super-resolution via contrastive representation learning, arXiv preprint arXiv:2107.00708 (2021).

\bibitem{bengio2013representation}
Y.~Bengio, A.~Courville, P.~Vincent, Representation learning: A review and new perspectives, IEEE Transactions on Pattern Analysis and Machine Intelligence 35~(8) (2013) 1798--1828.

\bibitem{dai2017deformable}
J.~Dai, H.~Qi, Y.~Xiong, Y.~Li, G.~Zhang, H.~Hu, Y.~Wei, Deformable convolutional networks, in: Proceedings of IEEE International Conference on Computer Vision, 2017, pp. 764--773.

\bibitem{dong2020multi}
H.~Dong, J.~Pan, L.~Xiang, Z.~Hu, X.~Zhang, F.~Wang, M.-H. Yang, Multi-scale boosted dehazing network with dense feature fusion, in: Proceedings of IEEE Conference on Computer Vision and Pattern Recognition, 2020, pp. 2157--2167.

\bibitem{simonyan2014very}
K.~Simonyan, A.~Zisserman, Very deep convolutional networks for large-scale image recognition, in: Proceedings of International Conference on Learning Representations, 2015.

\bibitem{deng2009imagenet}
J.~Deng, W.~Dong, R.~Socher, L.-J. Li, K.~Li, L.~Fei-Fei, Imagenet: A large-scale hierarchical image database, in: Proceedings of IEEE Conference on Computer Vision and Pattern Recognition, 2009, pp. 248--255.

\bibitem{charbonnier1994two}
P.~Charbonnier, L.~Blanc-Feraud, G.~Aubert, M.~Barlaud, Two deterministic half-quadratic regularization algorithms for computed imaging, in: Proceedings of IEEE International Conference on Image Processing, 1994, pp. 168--172.

\bibitem{ancuti2019dense}
C.~O. Ancuti, C.~Ancuti, M.~Sbert, R.~Timofte, Dense-haze: A benchmark for image dehazing with dense-haze and haze-free images, in: Proceedings of IEEE International Conference on Image Processing, 2019, pp. 1014--1018.

\bibitem{fu2017clearing}
X.~Fu, J.~Huang, X.~Ding, Y.~Liao, J.~Paisley, Clearing the skies: A deep network architecture for single-image rain removal, IEEE Transactions on Image Processing 26~(6) (2017) 2944--2956.

\bibitem{wei2019semi}
W.~Wei, D.~Meng, Q.~Zhao, Z.~Xu, Y.~Wu, Semi-supervised transfer learning for image rain removal, in: Proceedings of IEEE Conference on Computer Vision and Pattern Recognition, 2019, pp. 3877--3886.

\bibitem{zhang2018density}
H.~Zhang, V.~M. Patel, Density-aware single image de-raining using a multi-stream dense network, in: Proceedings of IEEE Conference on Computer Vision and Pattern Recognition, 2018, pp. 695--704.

\bibitem{yasarla2019uncertainty}
R.~Yasarla, V.~M. Patel, Uncertainty guided multi-scale residual learning-using a cycle spinning cnn for single image de-raining, in: Proceedings of IEEE Conference on Computer Vision and Pattern Recognition, 2019, pp. 8405--8414.

\bibitem{li2018recurrent}
X.~Li, J.~Wu, Z.~Lin, H.~Liu, H.~Zha, Recurrent squeeze-and-excitation context aggregation net for single image deraining, in: Proceedings of European Conference on Computer Vision, 2018, pp. 254--269.

\bibitem{ren2019progressive}
D.~Ren, W.~Zuo, Q.~Hu, P.~Zhu, D.~Meng, Progressive image deraining networks: A better and simpler baseline, in: Proceedings of IEEE Conference on Computer Vision and Pattern Recognition, 2019, pp. 3937--3946.

\bibitem{jiang2020multi}
K.~Jiang, Z.~Wang, P.~Yi, C.~Chen, B.~Huang, Y.~Luo, J.~Ma, J.~Jiang, Multi-scale progressive fusion network for single image deraining, in: Proceedings of IEEE Conference on Computer Vision and Pattern Recognition, 2020, pp. 8346--8355.

\bibitem{yang2017deep}
W.~Yang, R.~T. Tan, J.~Feng, J.~Liu, Z.~Guo, S.~Yan, Deep joint rain detection and removal from a single image, in: Proceedings of IEEE Conference on Computer Vision and Pattern Recognition, 2017, pp. 1357--1366.

\bibitem{zhang2014nighttime}
J.~Zhang, Y.~Cao, Z.~Wang, Nighttime haze removal based on a new imaging model, in: Proceedings of IEEE International Conference on Image Processing, 2014, pp. 4557--4561.

\bibitem{li2015nighttime}
Y.~Li, R.~T. Tan, M.~S. Brown, Nighttime haze removal with glow and multiple light colors, in: Proceedings of the IEEE International Conference on Computer Vision, 2015, pp. 226--234.

\bibitem{zhang2017fast}
J.~Zhang, Y.~Cao, S.~Fang, Y.~Kang, C.~Wen~Chen, Fast haze removal for nighttime image using maximum reflectance prior, in: Proceedings of IEEE Conference on Computer Vision and Pattern Recognition, 2017, pp. 7418--7426.

\bibitem{zhang2020nighttime}
J.~Zhang, Y.~Cao, Z.-J. Zha, D.~Tao, Nighttime dehazing with a synthetic benchmark, in: Proceedings of ACM International Conference on Multimedia, 2020, pp. 2355--2363.

\bibitem{zhang2019image}
H.~Zhang, V.~Sindagi, V.~M. Patel, Image de-raining using a conditional generative adversarial network, IEEE Transactions on Circuits and Systems for Video Technology 30~(11) (2019) 3943--3956.

\end{thebibliography}

\end{document}